\documentclass[10pt]{article}
\usepackage[accepted]{tmlr} 

\def\month{10}  
\def\year{2026} 
\def\openreview{\url{https://openreview.net/forum?id=VrWIwB8g3Z}}

\usepackage{amsmath}
\usepackage{amssymb}           
\usepackage{mathtools}         
\usepackage{mathrsfs} 
\usepackage{graphicx}      
\usepackage{subcaption}         
\usepackage{url}                
\usepackage{amsthm} 
\usepackage{thmtools}
\usepackage{algorithm, algpseudocode}
\usepackage{bbm}
\usepackage{booktabs}
\usepackage{enumitem}
\usepackage{microtype}

\usepackage[pagebackref=false,breaklinks=true,colorlinks=true,linkcolor=blue,citecolor=blue,urlcolor=blue]{hyperref}

\newtheorem{lem}{Lemma}
\newtheorem{defn}{Definition}

\newcommand{\beqan}{\begin{eqnarray*}}
\newcommand{\eeqan}{\end{eqnarray*}}

\newcommand{\beqa}{\begin{eqnarray}}
\newcommand{\eeqa}{\end{eqnarray}}

\newcommand{\Dir}{\mathrm{Dir}}
\newcommand{\DP}{\mathrm{DP}}

\title{Bayesian Neural Networks via Dirichlet Process Priors on the Data-Generating Distribution for Randomized Value Functions in Deep Reinforcement Learning}

\author{\name Sumit Vashishtha \email vash.sumit@gmail.com \\
      \addr Independent Researcher
    }

\begin{document}

{\maketitle} 

\begin{abstract}
Bayesian neural networks (BNNs) offer a principled framework for quantifying epistemic uncertainty. However, standard weight-space priors yield complex, high-dimensional neural network posteriors that are computationally prohibitive to sample from, and popular approximate inference schemes often fail to capture epistemic uncertainty effectively---a failure mode particularly acute in reinforcement learning (RL) \citep{osband2018randomized}. To resolve this challenge, we introduce a new class of BNNs that quantify uncertainty by placing Bayesian nonparametric Dirichlet Process (DP) priors directly on the underlying data-generating distribution rather than on network parameters. Building on this foundation, we develop DP-DQN, a reinforcement learning algorithm that leverages DP-BNNs to maintain posterior distributions over value functions and perform efficient Thompson sampling. By grounding randomized value functions in Bayesian nonparametrics, DP-DQN unifies data-driven posterior reweighting (via the Bayesian Bootstrap) with generative prior hypotheses injected directly through the DP base measure, enabling decisive ``deep exploration'' with a single neural network without the overhead of multi-network ensembles.
\end{abstract}

\section{Introduction}

In reinforcement learning (RL), maximizing cumulative return requires navigating the fundamental exploration-exploitation dilemma: balancing current best estimates (exploitation) with exploratory actions that resolve epistemic uncertainty. Solving this trade-off efficiently is challenging because delayed rewards require actions to be coordinated across extended multi-step decision sequences—a property known as \emph{deep exploration}~\citep{osband2016deep, osband2019behaviour, oren2025epistemic}—rather than perturbed independently at each individual state.

\textit{Thompson Sampling} \citep{thompson1933likelihood, kaufmann2012thompson, russo2018tutorial, durand2018streaming, riquelme2018deep, maillard2019mathematics} provides an elegant approach that tackles the exploration-exploitation dilemma by maintaining a posterior distribution over models and selecting actions in proportion to the probability that they are optimal. While these models are conventionally drawn from parametric distribution families (e.g., exponential families), Thompson sampling can in principle be applied to any statistical model admitting tractable posterior sampling \citep{riquelme2018deep}. A natural choice for such statistical models is Bayesian Neural Networks (BNNs), owing to their universal function approximation and rich representation learning capabilities. In contrast to standard neural networks that provide only point estimates, conventional BNNs specify a prior distribution over parameters $\theta$. Given $n$ training observations $z_{1:n}$, one infers a posterior distribution over $\theta$, which in turn captures epistemic uncertainty in function predictions.

In deep RL, one can represent value functions via BNNs and drive exploration via Thompson sampling over value-function predictions. However, exact posterior inference over neural network weights is notoriously intractable, prompting extensive research into approximate Bayesian methods. These range from Monte-Carlo (MC) dropout \citep{gal2016dropout} and variational inference \citep{blundell2015weight} to stochastic mini-batch Markov Chain Monte Carlo \citep{welling2011bayesian}. Unfortunately, deep RL algorithms built on these weight-space approximate Bayesian methods struggle on hard-exploration tasks because they fail to capture function-space epistemic uncertainty faithfully \citep{osband2018randomized}. This deficiency stems largely from the difficulty of designing meaningful priors directly in parameter space \citep{arbel2023primer}: while practitioners frequently possess domain knowledge regarding the data-generating process (e.g., transition dynamics and reward scales), mapping such insight into high-dimensional, isotropic parameter priors on over-parameterized neural weight vectors $\theta \in \mathbb{R}^d$ ($d \gg 1$) is rarely straightforward and often causes posterior uncertainty collapse.

To resolve these limitations, we introduce Bayesian Neural Networks via Dirichlet Process priors (DP-BNNs), a paradigm that quantifies epistemic uncertainty by placing a nonparametric DP prior directly on the underlying data-generating distribution rather than on network parameters. Under this formulation, posterior inference over the data distribution decomposes into two complementary components \citep{ferguson1973bayesian, vashishtha2025leveraging, vashishtha2026bayesian}: an empirical reweighting of observed data (the Bayesian Bootstrap) and an explicit prior base measure. In reinforcement learning, optimizing an action-value network on random measures sampled from this DP posterior induces a principled functional posterior distribution over value functions. Crucially, whereas existing randomized value function approaches require training an ensemble of multiple neural networks to represent functional uncertainty \citep{osband2016deep, osband2018randomized}, our approach achieves temporally coordinated deep exploration with a single neural network by freezing the value function sampled for each episode across the entire rollout.

We summarize our primary contributions as follows:
\begin{itemize}
    \item \textbf{DP-BNN Formulation and Sampling:} We introduce Bayesian Neural Networks via Dirichlet Process priors (DP-BNNs), a framework that quantifies epistemic uncertainty by placing a nonparametric DP prior on the data-generating distribution rather than on network weights. Using the Sethuraman stick-breaking construction, we develop a tractable posterior sampling algorithm for deep architectures and validate its epistemic uncertainty quantification on spatial 2D regression.
    
    \item \textbf{Deep Reinforcement Learning via DP-BNNs (DP-DQN):} We instantiate DP-BNNs for randomized value functions in deep Q-learning, establishing the DP-DQN algorithm. By optimizing an action-value DP-BNN on random measures sampled from the DP posterior via episodic boundary warmstarts, DP-DQN realizes principled Thompson sampling with exact multi-step temporal consistency using a single neural network.
    
    \item \textbf{Comprehensive Empirical Benchmarking:} We evaluate DP-BNNs across diverse decision-making benchmarks, demonstrating that they resolve hard exploration in DeepSea across extensive horizons evaluated up to $N=50$ ($2^{50} \approx 10^{15}$ trajectory search space with empirical $\mathcal{O}(N^{1.45})$ sample complexity), RiverSwim-6, deceptive $N$-Chain, continuous MountainCar-v0, and contextual Wheel Bandit, matching or outperforming multi-network ensembles while strictly maintaining a single-network footprint.
\end{itemize}

The remainder of this paper is organized as follows. Section~\ref{sec:related-works} surveys related work. Section~\ref{sec:Background-DP} reviews Dirichlet Processes and their stick-breaking construction. Section~\ref{sec:DP-BNNs} introduces Bayesian Neural Networks via Dirichlet Process priors (DP-BNNs) and details their posterior sampling algorithm. Section~\ref{sec:DP-BNN-randomized-value-functions} presents DP-DQN, applying DP-BNNs to randomized value functions in deep reinforcement learning, and evaluates its performance on challenging exploration benchmarks. Finally, Section~\ref{sec:conclusions} concludes the paper.

\section{Related Works}
\label{sec:related-works}

\paragraph{Randomized Value Functions and Exploration in Deep RL.}
Our work connects directly to deep reinforcement learning algorithms that explore via randomized value functions, replacing heuristic $\epsilon$-greedy exploration with principled posterior sampling (Thompson Sampling) \citep{thompson1933likelihood, kaufmann2012thompson, ghavamzadeh2015bayesian, osband2016deep, russo2018tutorial}. In deep RL, the most direct approach to Thompson sampling is to maintain a Bayesian neural network over action-values. However, exact posterior inference over network weights is intractable \citep{mackay2003information}, and mini-batch MCMC schemes scale poorly to online RL training regimes \citep{welling2011bayesian}. Consequently, practitioners have relied on approximate inference methods such as Monte-Carlo dropout \citep{gal2016dropout} and mean-field variational inference \citep{blundell2015weight}. Unfortunately, these weight-space approximations struggle to maintain functional uncertainty in unvisited regions, leading to exploration failure in challenging environments \citep{osband2018randomized}.

\paragraph{Bootstrap Ensembles and Prior Mechanisms.}
To approximate posterior sampling over function space without exact weight posteriors, \citet{lakshminarayanan2017simple} proposed Deep Ensembles, relying on random parameter initializations across independently trained networks. In sequential decision-making, \citet{osband2016deep} introduced Bootstrapped DQN, which approximates posterior sampling using an ensemble of $K$ neural networks trained on nonparametrically resampled data (see also \citealt{kveton2019garbage} for bootstrapping exploration in bandits). In sparse-reward environments, however, empirical bootstrapping alone provides limited guidance in regions devoid of data: because all ensemble heads are trained on unrewarded transitions, their value estimates tend to align across unvisited states. To address this and sustain directed exploration, \citet{osband2018randomized} introduced Bootstrapped DQN with Randomized Prior Functions (BootDQN-RP), which augments each trainable head $f_{\theta_k}$ with a fixed, randomly initialized prior network $p_k$. While highly effective in practice, this formulation substantially increases computational and memory demands, as it requires training and maintaining an ensemble of $K$ distinct neural networks alongside their corresponding prior networks. In another approach, Bayesian Deep Q-Networks (BDQN) \citep{azizzadenesheli2018efficient} applies Bayesian linear regression to the final layer representations; however, its capability on deep-exploration benchmarks has remained largely unexplored in the literature, and our empirical evaluations on DeepSea show that BDQN does not resolve the task. Theoretically, restricting Bayesian inference strictly to the final linear readout cannot account for epistemic uncertainty in the underlying feature representations on unvisited states (see Appendix~\ref{sec:why-bdqn-fails} for a detailed theoretical analysis).

\paragraph{Our Approach: Dirichlet Process Priors on Data Distributions.}
Rather than placing priors on network parameters \citep{blundell2015weight} or relying on multi-network ensembles with additive priors \citep{osband2018randomized}, our approach formulates epistemic uncertainty nonparametrically on the data-generating distribution. Building upon DP posterior sampling techniques developed in \citet{vashishtha2025leveraging} and \citet{vashishtha2026bayesian}, DP-BNNs connect Rubin's classical Bayesian Bootstrap \citep{rubin1981bayesian}---recovered in the non-informative limit $\alpha \to 0$---with an informative prior base measure $F_0$. Crucially, whereas empirical bootstrapping can only resample observed data, the base measure populates unvisited regions of the domain, sustaining non-vanishing epistemic uncertainty where experience is absent. This enables directed exploration with a single neural network while avoiding the parameter and memory footprint of multi-network ensembles.

\section{Background on Dirichlet Processes}
\label{sec:Background-DP}

This section reviews essential foundational concepts regarding Dirichlet Processes, conjugate posteriors, the Bayesian Bootstrap, and stick-breaking representations that underpin our DP-BNN framework; for comprehensive treatises on Bayesian nonparametric statistics, see \citet{ghosal2017fundamentals} and \citet{castillo2024bayesian}.

\paragraph{Dirichlet distribution.}
The Dirichlet distribution is a multivariate generalization of the Beta distribution. We denote the Dirichlet distribution parameterized by $(\alpha_{1}, \dots, \alpha_{n})$ on the probability simplex $\Delta^{n-1} = \{\mathbf{w} \in [0, 1]^n : \sum_{i=1}^n w_i = 1\}$ as $\Dir(\alpha_{1}, \dots, \alpha_{n})$, with density $p(\mathbf{w}) \propto \prod_{i=1}^{n}w_{i}^{\alpha_i-1}$.

\paragraph{Dirichlet Processes.}
In the Bayesian formalism, an unknown object of interest is treated as a random variable endowed with a prior distribution. When the unknown object is itself a probability measure on a measurable space $(\Theta, \mathcal{B}(\Theta))$ (e.g., a Polish space endowed with its Borel $\sigma$-algebra), defining a prior over infinite-dimensional distributions while satisfying unit-measure normalization is challenging. \citet{ferguson1973bayesian} introduced the \emph{Dirichlet Process} (DP)---a distribution over probability measures that induces finite-dimensional Dirichlet distributions on partitions. Concretely, a random probability measure $G$ is Dirichlet Process distributed with base probability measure $F_0$ and concentration parameter $\alpha \in \mathbb{R}^{+}$, denoted $G \sim \DP(\alpha, F_0)$, if for every finite measurable partition $(A_1, \dots, A_r)$ of $\Theta$:
\begin{equation}
(G(A_1), \dots, G(A_r)) \sim \Dir(\alpha F_0(A_1), \dots, \alpha F_0(A_r)).
\end{equation}

\paragraph{Conjugacy and Posteriors.}
A particularly useful property of the DP is its conjugacy: if observations $X_1, \dots, X_n \mid G \stackrel{\text{i.i.d.}}{\sim} G$ are drawn from an unknown distribution with prior $G \sim \DP(\alpha, F_0)$, the posterior distribution given $\mathbf{X}_n = (X_1, \dots, X_n)$ is itself a DP \citep{ferguson1973bayesian}:
\begin{equation} \label{def:DP-posteriors}
G \mid \mathbf{X}_n \sim \DP(\alpha_n, \overline{F}_n), \quad \text{where} \quad \alpha_n = \alpha + n, \quad \overline{F}_n = \frac{n}{\alpha+n}{F}_n + \frac{\alpha}{\alpha+n}F_0,
\end{equation}
where $F_n = \frac{1}{n}\sum_{i=1}^{n}\delta_{X_i}$ denotes the empirical measure of the observations \citep{ghosal2010dirichlet}.

Notice how the posterior base measure $\overline{F}_n$ in Equation~\eqref{def:DP-posteriors} naturally balances empirical evidence $F_n$ with prior structure $F_0$. As sample size $n \to \infty$, the empirical data dominates; conversely, when data is scarce ($n \to 0$), the prior $F_0$ governs. In our setting, $F_0$ serves as the foundational mechanism through which prior inductive biases enter the posterior; more broadly, recent advances in Bayesian nonparametrics demonstrate that random probability measures can also be elicited directly from prior information on meaningful functionals~\citep{gaffi2025random}.

\paragraph{Bayesian Bootstrap.}
The Bayesian Bootstrap \citep{rubin1981bayesian} provides a Bayesian analogue to Efron's classical bootstrap \citep{efron1994introduction}, and has been leveraged for randomized exploration in multi-armed bandits~\citep{kveton2019garbage}. Crucially, the Bayesian Bootstrap arises as the non-informative limiting case of a DP posterior as the prior concentration vanishes, $\alpha \to 0$ \citep{ghosal2017fundamentals}:
\begin{equation}\label{def:Bayesian-Bootstrap}
\mathrm{BB}_n := \lim_{\alpha \to 0} \DP\left(\alpha + n, \, \frac{n}{\alpha+n}F_n + \frac{\alpha}{\alpha+n}F_0\right) = \sum_{i=1}^{n} W_{i}\delta_{X_{i}},
\end{equation}
where $\mathbf{W}_n = (W_{1}, \dots, W_{n}) \sim \Dir(1, \dots, 1)$ and $\delta_{X_i}$ denotes a Dirac measure at observation $X_i$. The posterior mean of any linear functional under the Bayesian Bootstrap evaluates simply as $\mu(\mathrm{BB}_n) = \sum_{i=1}^{n} W_{i} X_{i} = \langle \mathbf{W}_n, \mathbf{X}_n \rangle$; for a comprehensive survey on the distributional properties of means and linear functionals of random probability measures, see \citet{lijoi2009distributional}.

\paragraph{Stick-Breaking Construction.}
While Equation~\eqref{def:DP-posteriors} specifies the posterior distribution of $G$, practical computation requires a constructive method to draw sample measures from a DP. \citet{sethuraman1994constructive} established that almost surely discrete random probability measures from $\DP(\alpha, F_0)$ admit an explicit stick-breaking representation:
\begin{equation}\label{def:Stick-Breaking-priors}
Q(\cdot) = \sum_{i=1}^{\infty} q_{i}\delta_{Z_{i}}(\cdot), \quad Z_i \stackrel{\mathrm{iid}}{\sim} F_0,
\end{equation}
where weights $q_i$ are generated by successively breaking pieces from a unit stick:
\begin{align} 
q_{1} &= V_{1}, \quad q_{i} = V_{i}\prod_{j=1}^{i-1}(1-V_{j}) \quad (i \ge 2), \label{def:SB-DP} \\
V_{i} &\stackrel{\mathrm{iid}}{\sim} \mathrm{Beta}(1,\alpha), \quad Z_{i} \stackrel{\mathrm{iid}}{\sim} F_0. \label{def:SB-DP-3}
\end{align}
In practice, the infinite series is truncated at a finite level $T$ by setting $V_T = 1$, which ensures the remaining stick length $q_T = \prod_{j=1}^{T-1}(1-V_j) = 1 - \sum_{i=1}^{T-1} q_i$ is fully assigned to $Z_T$, guaranteeing $\sum_{i=1}^T q_i = 1$ \citep{ishwaran2001gibbs, muliere1998approximating}. Truncation bounds and approximation guarantees are detailed in Appendix~\ref{sec:stick-breaking-finite}.

\paragraph{Iterative Sampling from the DP Posterior.}
To sample an explicit realization from the posterior $\DP(\alpha_n, \overline{F}_n)$, we leverage the sequential stick-breaking recursion of Dirichlet Processes. Conditioning a DP on a single observation $X_i$ satisfies the recursive distributional relation \citep{ferguson1973bayesian, ghosal2017fundamentals}:
\begin{equation}\label{eq:single-step-posterior}
Q_{i}(\cdot) \stackrel{d}{=} V_{i}\delta_{X_{i}} + (1-V_{i})Q_{i-1}(\cdot), \quad V_{i} \sim \mathrm{Beta}(1, \alpha + i),
\end{equation}
where $\stackrel{d}{=}$ denotes equality in distribution. Building upon this relation, \citet{vashishtha2025leveraging} derived an explicit multi-step stick-breaking representation for sampling realized probability measures directly from the DP posterior: initializing with a prior draw $Q_0 \sim \DP(\alpha, F_0)$ via stick-breaking and unrolling recursion~\eqref{eq:single-step-posterior} across all $n$ observations yields
\begin{equation}\label{eq:multi-step-posterior}
Q_{n} \stackrel{d}{=} V_{n}\delta_{X_{n}} + \sum_{i=1}^{n-1}\left[V_{i}\prod_{j=i+1}^{n}(1-V_{j})\right]\delta_{X_{i}} + \left[\prod_{i=1}^{n}(1-V_{i})\right] Q_{0}.
\end{equation}
The scalar coefficient $W_{\text{prior}} \triangleq \prod_{i=1}^{n}(1-V_{i})$ represents the total posterior probability mass retained by the prior measure $Q_0$. Under the sequential Beta stick-breaking construction where $1 - V_i \sim \mathrm{Beta}(\alpha + i, 1)$, this remaining prior weight follows the exact closed-form distribution:
\begin{equation}\label{eq:w-prior-beta}
W_{\text{prior}} = \prod_{i=1}^{n}(1-V_{i}) \sim \mathrm{Beta}(\alpha + 1, n),
\end{equation}
with expected prior mass $\mathbb{E}[W_{\text{prior}}] = \frac{\alpha+1}{\alpha+1+n}$. This term cleanly governs the balance between empirical evidence and prior belief: when data is absent ($n=0$), $W_{\text{prior}} = 1$ and the posterior recovers the prior $Q_0$; as observations accumulate ($n \to \infty$), $W_{\text{prior}} \to 0$ and empirical data dominates.

\section{Bayesian Neural Networks via DP Priors}
\label{sec:DP-BNNs}

In this section, we present a nonparametric methodology for sampling neural-network models from their posteriors to capture epistemic uncertainty in function predictions. The core idea is that predictive uncertainty originates from incomplete knowledge of the true data-generating distribution. Consider a neural network parameterized by $\theta$ trained under loss $\mathcal{L}_{\theta}(\cdot)$. If the true data-generating distribution $G^*(\cdot)$ were known (generating observations $\{z_i = (x_i, y_i)\}_{i=1}^n$, with inputs $x_i$ and targets $y_i$), optimal parameters $\theta^*$ would be obtained by minimizing the expected risk:
\begin{equation}\label{Eq:convoluted-loss}
\theta^* = \arg \min_{\theta} \mathbb{E}_{z \sim G^*}[\mathcal{L}_{\theta}(z)] = \arg \min_{\theta} \int \mathcal{L}_{\theta}(z) \, dG^*(z).
\end{equation}
In practice, however, the learner only observes a finite training set, leaving $G^*$ unknown. Epistemic uncertainty over $\theta$, and consequently over network predictions, is therefore a direct consequence of uncertainty over $G^*$.

To quantify this uncertainty, we place a Dirichlet Process prior directly on the unknown data distribution: $G \sim \mathrm{DP}(\alpha, F_0)$, where the base measure $F_0$ encapsulates prior domain structure and support (Lemma~\ref{thm:Support-of-DPs}). As training observations accumulate ($n \to \infty$), posterior consistency ensures that the posterior over $G$ concentrates weakly around the true distribution $G^*$ \citep{ferguson1973bayesian, ghosal2010dirichlet, castillo2024bayesian}. In finite-sample regimes, however—and particularly in regions where data is sparse or absent—the posterior does not collapse to an overconfident point estimate; instead, it retains non-vanishing epistemic variance governed by $F_0$. Whereas weight-space variational BNNs frequently underestimate uncertainty in voids or between separated data clusters~\citep{foong2019inbetween}, placing the nonparametric prior directly in data space preserves functional uncertainty wherever data has not yet been observed.

Given a finite training dataset $\mathcal{D}_n = \{z_i\}_{i=1}^n$, the posterior over the data-generating distribution is $G^* \mid \mathcal{D}_n \sim \DP(\alpha_n, \overline{F}_n)$ with posterior indices defined in Equation~\eqref{def:DP-posteriors}. Applying the stick-breaking construction (Equations~\eqref{def:SB-DP}--\eqref{def:SB-DP-3}) and the iterative posterior recursion (Equation~\eqref{eq:multi-step-posterior})~\citep{vashishtha2025leveraging}, any random posterior measure $P_j \sim \DP(\alpha_n, \overline{F}_n)$ truncated at $T$ prior atoms takes the explicit finite discrete form:
\begin{equation}\label{Eq:Stick-Breaking-priors-Ch4-Pj}
P_j(\cdot) = \sum_{i=1}^{n+T} p'_i \delta_{z_i}(\cdot),
\end{equation}
where $\{z_i\}_{i=1}^n$ are the observed training points and $\{z_{n+t}\}_{t=1}^T \stackrel{\mathrm{iid}}{\sim} F_0$ are prior pseudo-observations sampled from the base measure. The weights $\{p'_i\}_{i=1}^{n+T}$ satisfy $\sum_{i=1}^{n+T} p'_i = 1$, where the empirical weights $\{p'_i\}_{i=1}^n$ are obtained from Equation~\eqref{eq:multi-step-posterior} and the prior weights $\{p'_{n+t}\}_{t=1}^T$ are generated by scaling stick-breaking weights from Equation~\eqref{def:SB-DP} by the remaining prior factor $W_{\text{prior}} = \prod_{k=1}^n (1 - V_k)$ (Eq.~\eqref{eq:w-prior-beta}).

Substituting this discrete measure into the risk functional in Equation~\eqref{Eq:convoluted-loss}, sampling a neural-network model $\theta^{(j)}$ from the posterior reduces directly to a weighted empirical loss:
\begin{equation}\label{eq:convoluted-loss-Pj}
\theta^{(j)} = \arg \min_{\theta} \sum_{i=1}^{n+T} p'_i \mathcal{L}_{\theta}(z_i).
\end{equation}
The complete procedure is summarized in Algorithm~\ref{alg:DP-NN}. In order to compute variations in predictions and confidence intervals, one computes predictions across the sampled models $\{\theta^{(j)}\}_{j=1}^J$. Notice that as $\alpha \to 0$, $P_j$ recovers the Bayesian Bootstrap (Equation~\eqref{def:Bayesian-Bootstrap}), placing probability mass solely on observed data points $\{z_i\}_{i=1}^n$. When $\alpha > 0$, the DP base measure $F_0$ incorporates prior domain structure and ensures non-vanishing epistemic uncertainty in unexplored regions of the input space.

\begin{algorithm}[t]
\caption{Posterior Sampling of Bayesian Neural Networks via Dirichlet Process Priors (DP-BNN)}
\label{alg:DP-NN}
\begin{algorithmic}[1]
\State \textbf{Input:} Training observations $\mathcal{D}_n = \{z_i\}_{i=1}^n$, DP prior parameters $(\alpha, F_0)$, truncation level $T$, sample count $J$, loss function $\mathcal{L}_\theta(\cdot)$
\State \textbf{Output:} Posterior model samples $\{\theta^{(j)}\}_{j=1}^J$
\For{$j = 1, \dots, J$}
    \State Sample synthetic prior pseudo-observations: $\{z_{n+t}\}_{t=1}^T \stackrel{\text{i.i.d.}}{\sim} F_0$
    \State Sample stick-breaking weights $\{p'_i\}_{i=1}^{n+T} \in \Delta^{n+T-1}$ via Equations~\eqref{Eq:Stick-Breaking-priors-Ch4-Pj} and \eqref{eq:multi-step-posterior}
    \State Form random discrete posterior measure: $P_j = \sum_{i=1}^{n+T} p'_i \delta_{z_i}$
    \State Initialize network parameters $\theta^{(j)}$
    \State Optimize parameters to convergence: $\theta^{(j)} \leftarrow \arg\min_\theta \sum_{i=1}^{n+T} p'_i \mathcal{L}_\theta(z_i)$ via Eq.~\eqref{eq:convoluted-loss-Pj}
\EndFor
\State \Return $\{\theta^{(j)}\}_{j=1}^J$
\end{algorithmic}
\end{algorithm}

\begin{figure}[!htbp]
\centering
\includegraphics[width=\linewidth]{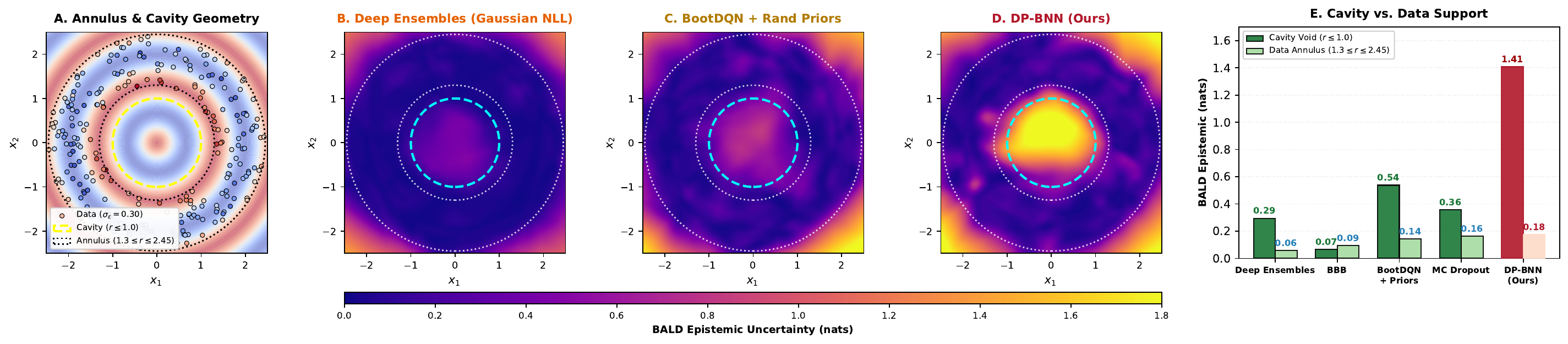}
\caption{\textbf{Epistemic uncertainty on 2D annular regression.} \textbf{(A)} Annular training data ($r \in [1.3, 2.45]$) with unobserved central cavity ($r \le 1.0$). \textbf{(B--D)} BALD epistemic uncertainty heatmaps (nats)~\citep{houlsby2011bayesian} on a shared $[0.0, 1.8]$ scale for Deep Ensembles (B), BootDQN with randomized priors (C), and DP-BNN (D). \textbf{(E)} Void vs.\ data support BALD contrast across evaluated methods.}
\label{fig:2D-regression-uncertainty}
\end{figure}

We evaluate the epistemic uncertainty of DP-BNNs on a 2D regression benchmark over an annular manifold where 2D spatial inputs $x = (x_1, x_2) \in [-2.5, 2.5]^2$ have radial distance $r = \|x\|_2 \in [1.3, 2.45]$ with noisy scalar target observations $y = \cos(1.5\pi r) + \epsilon \in [-1, 1]$ ($\sigma_{\rm aleatoric} = 0.30$), leaving an unobserved central cavity void for $r \le 1.0$ (Figure~\ref{fig:2D-regression-uncertainty}A). Because there are no training observations within the interior cavity, a reliable Bayesian model is fundamentally expected to exhibit high epistemic uncertainty (low predictive confidence) throughout this unobserved void, while contracting with high confidence on the data-rich annular ring. We quantify epistemic uncertainty via Bayesian Active Learning by Disagreement (BALD)~\citep{houlsby2011bayesian} across $J=10$ sampled models (Appendix~\ref{sec:bald-details}). Crucially, unlike deep ensembles~\citep{lakshminarayanan2017simple} that rely on disparate parameter initializations to generate variance, DP-BNN induces epistemic diversity non-parametrically in function space via synthetic pseudo-observations from a product base measure $F_0(x, y) = F_{0x}(x) \times F_{0y}(y)$. Following Lemma~\ref{thm:Support-of-DPs}, we choose the support of $F_{0x}$ as the uniform distribution over the domain $[-2.5, 2.5]^2$ to ensure non-empty coverage of unobserved voids, paired with dispersed prior targets $F_{0y} = \mathcal{N}(0, 1.5^2)$ ($T=50$ atoms, concentration parameter $\alpha=10.0$). As shown in Figure~\ref{fig:2D-regression-uncertainty}B--E, DP-BNN reflects this expected epistemic behavior, forming an isotropic uncertainty region in the void ($1.41$ nats vs.\ $0.18$ nats on data support; $8.0\times$ contrast ratio). In comparison, standard baselines exhibit lower contrast between the void and data support: BootDQN with randomized priors~\citep{osband2018randomized} yields $0.54$ vs.\ $0.14$ nats ($3.8\times$), MC Dropout~\citep{gal2016dropout} yields $0.36$ vs.\ $0.16$ nats ($2.2\times$), Deep Ensembles~\citep{lakshminarayanan2017simple} yields $0.29$ vs.\ $0.06$ nats ($4.8\times$), and variational inference (Bayes By Backprop, BBB)~\citep{blundell2015weight} produces near-uniform uncertainty across the domain ($0.07$ vs.\ $0.09$ nats). Because each model $\theta^{(j)}$ minimizes the loss under an independently drawn measure $P_j$, the training cost of DP-BNN is on par with standard ensemble baselines (wall-clock training times for $J=10$ models: DP-BNN [$5.9\,$s], BootDQN+priors [$5.4\,$s], Deep Ensembles [$6.9\,$s], BBB [$2.1\,$s], MC Dropout [$0.7\,$s]).

While quantifying predictive variance in this benchmark required training multiple models ($J=10$), Thompson sampling in sequential decision-making requires only a single sampled model per episode \citep{thompson1933likelihood, russo2018tutorial}. Because DP-BNNs represent uncertainty via the DP posterior over data distributions, training a single neural network on a draw from this posterior directly provides the required sample. Consequently, an agent can perform Thompson sampling with a single neural network, eliminating the heuristic multi-network ensembles required by previous methods. In the next section, we apply this framework to deep reinforcement learning.

\section{Reinforcement Learning with DP-BNNs}
\label{sec:DP-BNN-randomized-value-functions}

We now extend the DP-BNN framework to reinforcement learning, developing DP-DQN: a deep Q-learning algorithm that achieves directed, temporally consistent exploration via posterior sampling over action-value functions.

\subsection{Problem Formulation and Exploration Bottlenecks}
\label{sec:drl-background}

\paragraph{Markov Decision Processes.}
A Markov Decision Process (MDP)~\citep{puterman1990markov} is formalized by the tuple $M = (\mathcal{S}, \mathcal{A}, R, P, \gamma)$, where $\mathcal{S}$ is the state space, $\mathcal{A}$ is the discrete action space, $R: \mathcal{S} \times \mathcal{A} \to \mathbb{R}$ is the reward function, $P(\cdot \mid s, a)$ is the transition probability distribution, and $\gamma \in [0, 1)$ is the discount factor. An agent aims to identify an optimal policy $\pi^*$ maximizing expected cumulative discounted return, characterized by the unique optimal action-value function $Q^*(s, a)$ satisfying the Bellman optimality equation:
\begin{equation}\label{Eq:Bellman-optimality}
Q^*(s, a) = R(s, a) + \gamma \mathbb{E}_{s' \sim P(\cdot \mid s, a)} \left[\max_{a' \in \mathcal{A}} Q^*(s', a')\right].
\end{equation}

\paragraph{Deep Q-Learning and Exploration Bottlenecks.}
In complex or continuous state spaces, tabular representation is infeasible. Value-based deep reinforcement learning methods such as Deep Q-Networks (DQN) and Double DQN~\citep{mnih2015human, van2016deep, bellemare2017distributional} approximate $Q^*(s, a)$ with a deep neural network $Q(s, a; \theta)$ parameterized by $\theta$. Parameters are optimized by minimizing the mean-squared Temporal Difference (TD) regression loss over transition tuples $z = (s, a, r, s', d)$ sampled from an experience replay buffer $\mathcal{B}$:
\begin{equation}\label{eq:DQN-loss}
\mathcal{L}_{\mathrm{DQN}}(\theta; \theta^-) = \mathbb{E}_{z \sim \mathcal{B}}\left[\ell_{\mathrm{TD}}(z; \theta, \theta^-)\right], \quad \text{where} \quad \ell_{\mathrm{TD}}(z; \theta, \theta^-) = \left( r + (1 - d)\gamma \max_{a' \in \mathcal{A}} Q(s', a'; \theta^-) - Q(s, a; \theta) \right)^2,
\end{equation}
where $\theta^-$ denotes the parameters of a tracking target network. In the framework of statistical learning theory, Eq.~\eqref{eq:DQN-loss} is the empirical risk minimizer of the expected TD loss $\mathbb{E}_{z \sim G^*}[\ell_{\mathrm{TD}}(z; \theta, \theta^-)]$ under the empirical measure $\mathbb{P}_n = \frac{1}{n} \sum_{i=1}^n \delta_{z_i}$ of the replay buffer, corresponding to neural Fitted Q-Iteration~\citep{ernst2005tree, riedmiller2005neural, mnih2015human}.

Standard DQN relies on heuristic $\epsilon$-greedy dithering for exploration. However, independent action perturbations at each step lack temporal consistency, requiring an exponential sample complexity $\Omega(|\mathcal{A}|^H)$ to discover delayed rewards over horizon $H$~\citep{osband2018randomized, osband2019behaviour}. Principled deep exploration instead requires Posterior Sampling for Reinforcement Learning (PSRL)~\citep{dearden1998bayesian, strens2000bayesian, osband2013more, ghavamzadeh2015bayesian}, where an agent samples an action-value hypothesis from an epistemic posterior and acts greedily with respect to it for the entire episode rollout. However, existing deep exploration approaches either maintain an ensemble of $K$ separate neural networks (e.g., Bootstrapped DQN; \citealt{osband2016deep}), which incurs a prohibitive $K$-fold memory and computational footprint, train parameterized epistemic hypermodels~\citep{dwaracherla2020hypermodels}, inject heuristic parameter noise (NoisyNet; \citealt{fortunato2018noisy}), or rely on parametric weight-space Bayesian approximations (e.g., Bayes By Backprop, MC Dropout) that suffer from prior generalization collapse and fail to sustain epistemic uncertainty across deep horizons~\citep{foong2019inbetween, osband2018randomized}. DP-BNNs resolve these dual bottlenecks—computational scaling overhead and epistemic uncertainty collapse—by placing a DP prior directly on the data-generating distribution of transition experience $G^*$, inducing rich function-space epistemic diversity nonparametrically with a single neural network (instantiated here as DP-DQN).

\subsection{DP-DQN: Posterior Sampling for Randomized Value Functions}
\label{sec:dp-dqn-formulation}

While standard DQN minimizes empirical TD loss over observed transitions in the replay buffer (Eq.~\eqref{eq:DQN-loss}), the DP-BNN framework quantifies epistemic uncertainty by placing a DP prior directly on the underlying transition-experience distribution:
\begin{equation}\label{eq:dp_prior_rl}
G^* \sim \mathrm{DP}(\alpha, F_0),
\end{equation}
where the base measure $F_0$ generates synthetic transition tuples $\tilde{z} = (\tilde{s}, \tilde{a}, \tilde{r}, \tilde{s}', \tilde{d})$ via factorized marginals $F_0(s, a, r, s', d) = F_{0s}(s) \times F_{0a}(a) \times F_{0r}(r) \times F_{0s'}(s') \times F_{0d}(d)$, with continuing transitions ($\tilde{d}=0$). Following Lemma~\ref{thm:Support-of-DPs}, the base measure maintains full topological support over state-action spaces ($\mathrm{supp}(F_{0s}) \supseteq \mathcal{S}$, $\mathrm{supp}(F_{0a}) = \mathcal{A}$). Because data collection in reinforcement learning is policy-dependent, the reward prior $F_{0r}$ is specified to sustain exploratory variance in unvisited regions (Appendix~\ref{sec:dp-dqn-base-choice}).

Given $n$ observed transitions in the replay buffer $\mathcal{B} = \{z_i = (s_i, a_i, r_i, s'_i, d_i)\}_{i=1}^n$, Ferguson's conjugacy~\citep{ferguson1973bayesian} yields the posterior distribution over the data-generating distribution:
\begin{equation}\label{eq:dp_posterior_rl}
G^* \mid \mathcal{B} \sim \mathrm{DP}\left(\alpha + n, \; \frac{n}{\alpha + n} \mathbb{P}_n + \frac{\alpha}{\alpha + n} F_0\right), \quad \text{where } \mathbb{P}_n = \frac{1}{n} \sum_{i=1}^n \delta_{z_i}.
\end{equation}
Under this formulation, posterior sampling over value functions corresponds to drawing a discrete random measure $P_k \sim G^* \mid \mathcal{B}$ for each episode $k$ and optimizing the action-value network under $P_k$: $\theta_k = \arg\min_\theta \mathbb{E}_{z \sim P_k}[\ell_{\mathrm{TD}}(z; \theta, \theta^-)]$.

\begin{figure}[t]
\centering
\includegraphics[width=\linewidth]{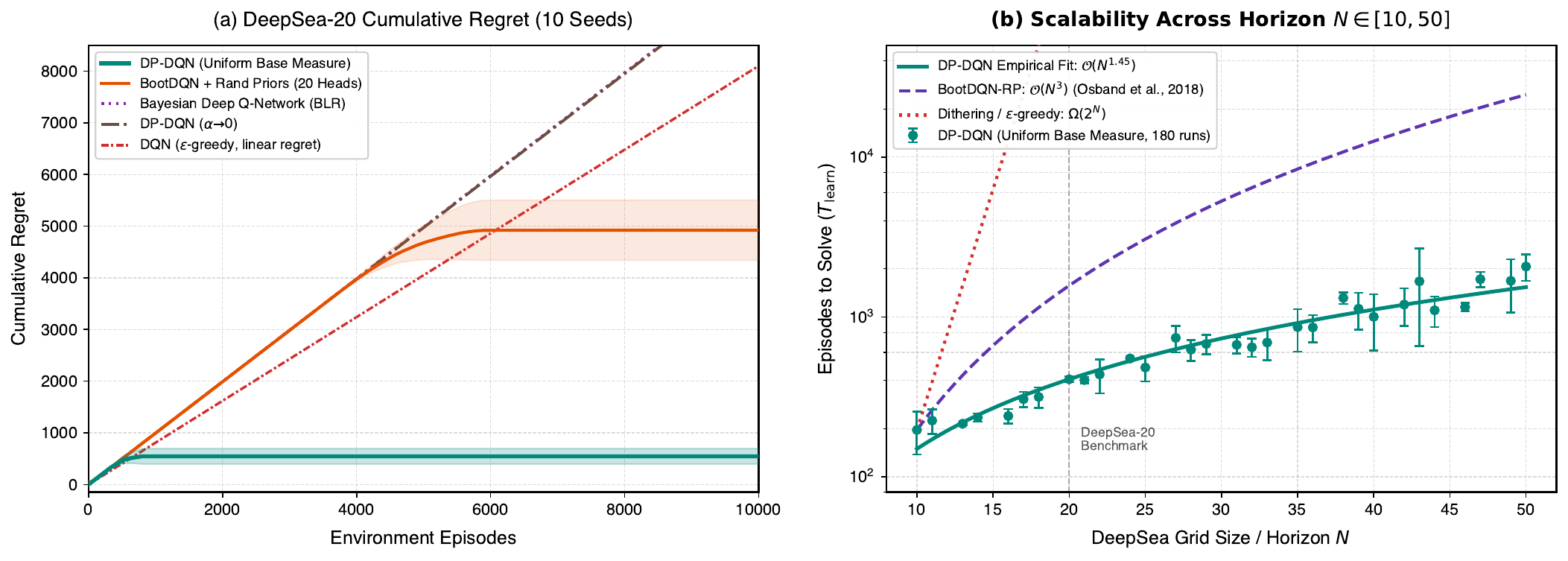}
\caption{\textbf{Empirical exploration and scaling on DeepSea}~\citep{osband2019behaviour}. \textbf{(a)} Cumulative regret on DeepSea-20 (10 random seeds, shaded bands denote $\pm 1$ standard deviation) comparing DP-DQN against BootDQN-RP, BDQN, and DP-DQN in the Bayesian bootstrap limit ($\alpha \to 0$; Section~\ref{sec:Background-DP}). \textbf{(b)} Empirical sample complexity across 30 grid sizes up to $N=50$ (6 random seeds per size, error bars denote $\pm 1$ standard deviation), showing empirical $\mathcal{O}(N^{1.45})$ scaling for DP-DQN compared to $\mathcal{O}(N^3)$ for BootDQN-RP.}
\label{fig:DP-DQN-regret}
\end{figure}

\paragraph{Full-Buffer Conditioning and Prior Swamping.}
However, directly realizing this posterior draw over the full replay buffer $\mathcal{B}$ introduces two practical challenges:
\begin{enumerate}[leftmargin=*,noitemsep,topsep=0pt]
\item \emph{Computational overhead:} Evaluating stick-breaking representations and backpropagating across $|\mathcal{B}| \sim 10^5$ stored transitions at every episode boundary is computationally prohibitive.
\item \emph{Prior swamping:} In sparse-reward exploration, unguided initial rollouts inevitably accumulate thousands of unrewarded transitions. If the sample size $n$ in Eq.~\eqref{eq:dp_posterior_rl} scales with the raw buffer size ($n \sim |\mathcal{B}| \sim 10^5$), the relative weight of the prior base measure $\frac{\alpha}{\alpha + n} \to 0$ rapidly diminishes. The posterior contracts toward empirical resampling over observed transitions—recovering the Bayesian Bootstrap limit ($\alpha \to 0$)~\citep{rubin1981bayesian}—which diminishes exploratory drive early in training (Figure~\ref{fig:DP-DQN-regret}a; Appendix~\ref{sec:stochastic-warmstart-theory-appendix}).
\end{enumerate}

\paragraph{Composite DP-Bellman Loss.}
To approximate this posterior without full-buffer computation, DP-DQN evaluates an empirical mini-batch $\{z_i\}_{i=1}^B \sim \mathcal{B}$ alongside $T$ synthetic prior atoms $\{\tilde{z}_j\}_{j=1}^T \sim F_0$, forming the discrete composite measure $P_k = \sum_{i=1}^B w_i^{\text{emp}} \delta_{z_i} + W_{\text{prior}} \sum_{j=1}^T q_j^{(0)} \delta_{\tilde{z}_j}$. Minimizing the expected temporal-difference error $\mathbb{E}_{z \sim P_k}[\ell_{\mathrm{TD}}(z; \theta, \theta^-)]$ yields the \emph{composite DP-Bellman loss}:
\begin{equation}
\label{eq:dp_bellman_loss}
\mathcal{L}_{\text{DP}}(\theta) = \sum_{i=1}^B w_i^{\text{emp}} \big(y_i - Q(s_i, a_i; \theta)\big)^2 + W_{\text{prior}} \sum_{j=1}^T q_j^{(0)} \big(\tilde{y}_j - Q(\tilde{s}_j, \tilde{a}_j; \theta)\big)^2,
\end{equation}
where $y_i = r_i + (1 - d_i)\gamma \max_{a'} Q(s'_i, a'; \theta^-)$ and $\tilde{y}_j = \tilde{r}_j + (1 - \tilde{d}_j)\gamma \max_{a'} Q(\tilde{s}'_j, a'; \theta^-)$ denote empirical and prior Bellman targets, and $\theta^-$ tracks target network parameters via Polyak averaging $\theta^- \leftarrow (1 - \tau) \theta^- + \tau \theta_k$~\citep{lillicrap2015continuous}. Layer Normalization~\citep{ba2016layer} is applied across hidden layers to ensure numerical stability across heterogeneous stick-breaking weights.

\paragraph{Information Calibration via Bellman Surprise.}
Scaling the prior weight $W_{\text{prior}} \sim \mathrm{Beta}(\alpha + 1, n)$ by the raw buffer size would cause the prior mass to vanish in early unrewarded rollouts (prior swamping). To prevent this, we modulate the prior mass by the cumulative information content of observed transitions rather than raw sample count. Setting $W_{\text{prior}} \sim \mathrm{Beta}(\alpha + 1, N_{\text{stat}})$, the effective sample counter $N_{\text{stat}}$ accumulates sample weight proportional to normalized Bellman surprise:
\begin{equation}
\label{eq:n_stat_update}
N_{\text{stat}} \leftarrow N_{\text{stat}} + \frac{1}{B} \sum_{i=1}^B \min\left(1, \; \frac{|\delta_i|}{\sigma_{\text{TD}}}\right), \quad \delta_i = y_i - Q(s_i, a_i; \theta),
\end{equation}
where $\sigma_{\text{TD}}$ denotes the running standard deviation of Bellman errors. Transitions with near-zero Bellman error represent well-modeled dynamics that contribute negligible information, maintaining non-vanishing prior weight until novel reward signals are uncovered (Appendix~\ref{sec:stochastic-warmstart-theory-appendix}).

\paragraph{Episodic Boundary Warmstart and Rollout Freezing.}
To preserve multi-step temporal consistency, DP-DQN eliminates intra-episode parameter updates, amortizing optimization into $W = \lceil H / 2 \rceil$ stochastic warmstart gradient steps at episode boundaries. At the start of episode $k$, prior atoms $\{\tilde{z}_j\}_{j=1}^T \sim F_0$ are drawn once to fix the episodic exploration hypothesis. Across steps $w = 1, \dots, W$, fresh empirical mini-batches $\mathcal{D}_w \sim \mathcal{B}$ are sampled to optimize $\theta$ on $\mathcal{L}_{\text{DP}}(\theta)$, which approximately draws a value function from the functional posterior. In accordance with PSRL \citep{strens2000bayesian, osband2013more}, the updated parameters $\theta_k \leftarrow \theta$ are frozen throughout the episode rollout, executing greedy actions $a_t = \arg\max_a Q(s_t, a; \theta_k)$ to guarantee temporally coordinated exploration with a single neural network (Algorithm~\ref{alg: deep-rl-dp}).

\begin{algorithm}[t]
\caption{DQN with Thompson Sampling-Based Exploration via DP-BNNs}
\label{alg: deep-rl-dp}
{\small
\begin{algorithmic}[1]
\State \textbf{Input:} Action-value network $Q(s, a; \theta)$, target network parameters $\theta^-$, DP prior parameters $(\alpha, F_0)$, truncation level $T$, replay buffer $\mathcal{B}$, mini-batch size $B$, warmstart steps $W$, learning rate $\eta$, discount factor $\gamma \in [0, 1)$, episode horizon $H$, Polyak rate $\tau$
\State Initialize effective information counter $N_{\text{stat}} \leftarrow 0$
\For{episode $k = 1, 2, \dots$}
    \State \textbf{Stage 1: Value Function Posterior Sampling via DP-BNNs (Warmstart)}
    \State Sample synthetic prior transitions: $\tilde{z}_j = (\tilde{s}_j, \tilde{a}_j, \tilde{r}_j, \tilde{s}'_j, \tilde{d}_j) \stackrel{\text{iid}}{\sim} F_0$ for $j = 1, \dots, T$
    \State Compute prior stick weights $q_j^{(0)}$ via Eq.~\eqref{def:SB-DP} for $j = 1, \dots, T$
    \For{step $w = 1, \dots, W$}
        \State Sample empirical mini-batch $\mathcal{D}_w = \{z_i = (s_i, a_i, r_i, s'_i, d_i)\}_{i=1}^B \sim \mathcal{B}$
        \State Sample prior factor $W_{\text{prior}} \sim \mathrm{Beta}(\alpha + 1, N_{\text{stat}})$ and empirical weights $w_i^{\text{emp}}$ via Eq.~\eqref{eq:multi-step-posterior}
        \State Compute Bellman targets: $y_i = r_i + (1 - d_i) \gamma \max_{a'} Q(s'_i, a'; \theta^-)$ and $\tilde{y}_j = \tilde{r}_j + (1 - \tilde{d}_j) \gamma \max_{a'} Q(\tilde{s}'_j, a'; \theta^-)$
        \State Update online network: $\theta \leftarrow \theta - \eta \nabla_\theta \mathcal{L}_{\text{DP}}(\theta)$ \quad (via Eq.~\eqref{eq:dp_bellman_loss})
        \State Calibrate information counter $N_{\text{stat}}$ via TD surprise (Eq.~\eqref{eq:n_stat_update})
    \EndFor
    \State Freeze episode policy network: $\theta_k \leftarrow \theta$
    \State \textbf{Stage 2: Temporally Coordinated Rollout under Sampled Hypothesis}
    \State Observe initial state $s_0$ from environment
    \For{$t = 0, \dots, H-1$}
        \State Select action greedily under sampled model: $a_t = \arg\max_a Q(s_t, a; \theta_k)$
        \State Execute $a_t$, observe reward $r_t$, next state $s_{t+1}$, and terminal flag $d_t$
        \State Store transition $z_t = (s_t, a_t, r_t, s_{t+1}, d_t)$ into replay buffer $\mathcal{B}$
        \If{$d_t$ is true} \textbf{break} \EndIf
    \EndFor
    \State Synchronize target network via Polyak tracking: $\theta^- \leftarrow (1 - \tau) \theta^- + \tau \theta_k$
\EndFor
\end{algorithmic}
}
\end{algorithm}

\subsection{Empirical Evaluation on Hard-Exploration Benchmarks}
\label{sec:empirical-deep-exploration}

To evaluate whether DP-BNN posteriors provide effective randomized value functions with a single neural network, we examine DP-DQN across the scalable DeepSea benchmark~\citep{osband2019behaviour}, the canonical diagnostic domain for deep exploration. Specifically, we evaluate: (i) whether a single neural network equipped with a DP posterior matches the sample efficiency of multi-network ensembles on canonical DeepSea-20; (ii) how empirical sample complexity scales as grid size expands up to $N=50$; and (iii) how the computational and memory footprint of DP-DQN compares with multi-network ensemble baselines.

\paragraph{The DeepSea Exploration Benchmark.}
DeepSea requires an agent to execute $N$ consecutive coordinated actions across an episode of length $N$ to collect a solitary terminal reward ($r = +1.0$) at $(N-1, N-1)$. It challenges exploration through three distinct mechanisms: (i) an unknown action permutation per row masks the optimal direction; (ii) a continuous step cost of $-0.01/N$ penalizes exploration along the optimal path and biases myopic policies toward the unrewarded boundary; and (iii) any suboptimal step irreversibly routes the agent into zero-reward absorbing states. Consequently, dithering exploration strategies (such as $\epsilon$-greedy) require an expected $\Omega(2^N)$ episodes to discover the goal~\citep{osband2016deep, osband2019behaviour}---scaling from $2^{20} \approx 10^6$ episodes at $N=20$ to $2^{50} \approx 10^{15}$ episodes at $N=50$. We benchmark on canonical DeepSea-20 (Figure~\ref{fig:DP-DQN-regret}a) and evaluate scaling behavior across 30 grid sizes up to $N=50$ (Figure~\ref{fig:DP-DQN-regret}b).

\paragraph{Base Measure for DP-DQN.}
To ensure exploration is driven without privileged environment knowledge, DP-DQN specifies an uninformative product base measure over transitions, $F_0(s, a, r, s', d) = F_{0s}(s) \times F_{0a}(a) \times F_{0r}(r) \times F_{0s'}(s') \times F_{0d}(d)$. State and next-state components are sampled uniformly over the $N^2$ canonical one-hot vectors ($\tilde{s}, \tilde{s}' \sim \mathrm{Uniform}(\mathcal{S}_{\mathrm{one\text{-}hot}} \subset \{0, 1\}^{N^2})$), actions uniformly $\tilde{a} \sim \mathrm{Uniform}(\{0, 1\})$, and continuing flags $\tilde{d}=0$, encoding zero information about transition graph topology or goal coordinates. For rewards, setting $F_{0r} = \mathcal{U}(0, R_{\max})$ (with $R_{\max} = 1.0$) naturally induces stochastic optimism \citep{osband2017posterior, agrawal2017optimistic}, sustaining positive expected value on unexplored states to overcome continuous step costs without artificial reward shaping. We set concentration parameter $\alpha = 3.0$ and truncation level $T = 50$; complete hyperparameter specifications, support conditions, and base measure ablations are detailed in Table~\ref{tab:hyperparameters} and Appendices~\ref{sec:dp-dqn-base-choice} and~\ref{sec:stochastic-optimism-and-bdqn-failure}.

\paragraph{DeepSea-20 Exploration Performance.}
Figure~\ref{fig:DP-DQN-regret}a presents comparative results on DeepSea-20 across 10 independent random seeds per algorithm over 10,000 episodes (mean cumulative regret with shaded $\pm 1$ standard deviation). All agents employ an action-value architecture with a single hidden layer of 20 units (MLP-20), matching the network capacity in~\citet{osband2018randomized}. For DP-DQN, Layer Normalization~\citep{ba2016layer} is included to stabilize gradient variance across heterogeneous stick-breaking weights, with concentration parameter $\alpha = 3.0$ and truncation level $T = 50$ (Table~\ref{tab:hyperparameters}). Equipped with the uninformative uniform base measure, DP-DQN achieves a 100\% solve rate (10/10 seeds), discovering the terminal goal at episode $397 \pm 154$ and reaching policy convergence in $602 \pm 104$ episodes, with cumulative regret plateauing at $418 \pm 94$. In comparison, Bootstrapped DQN with randomized priors (BootDQN-RP, 20 ensemble heads + 20 prior networks)~\citep{osband2018randomized} requires $5{,}210 \pm 604$ episodes with $4{,}922 \pm 582$ regret. Notably, DP-DQN consolidates into the optimal policy within $\approx 200$ episodes after initial discovery ($397 \pm 154 \to 602 \pm 104$). Because the high Bellman surprise of the newly discovered goal sharply increments the information counter $N_{\text{stat}}$, the prior factor $W_{\text{prior}}$ is rapidly attenuated in favor of empirical Bellman targets (Eq.~\eqref{eq:dp_bellman_loss}), stabilizing greedy execution without the inter-head variance or conflicting prior biases that can impede policy consolidation in heuristic ensembles. Conversely, standard DQN, BDQN~\citep{azizzadenesheli2018efficient}, and DP-DQN in the Bayesian bootstrap limit ($\alpha \to 0$; Section~\ref{sec:Background-DP}, Eq.~\eqref{def:Bayesian-Bootstrap}) fail to discover the goal across all seeds (0/10 seeds solved), incurring linear regret ($>9{,}900$). In this Bayesian bootstrap limit where the prior concentration parameter vanishes ($\alpha \to 0$), the agent samples purely from empirical replay transitions without synthetic prior augmentation ($W_{\text{prior}} = 0$). Because early trajectories observe only zero-reward transitions, pure empirical resampling provides no exploratory optimism toward unvisited states, confirming that the prior base measure $F_0$ is essential for driving deep exploration (Appendices~\ref{sec:stochastic-warmstart-theory-appendix} and~\ref{sec:why-bdqn-fails}).

\paragraph{Scalability Across DeepSea Grid Size \texorpdfstring{$N \in [10, 50]$}{N in [10, 50]}.}
To evaluate scalability as the exploration horizon expands, we benchmarked DP-DQN across 30 grid sizes up to $N=50$ (6 independent random seeds per size, error bars denote $\pm 1$ standard deviation; Figure~\ref{fig:DP-DQN-regret}b and Table~\ref{tab:empirical_summary}). To accommodate the expanding state space ($|\mathcal{S}| = 2{,}500$) and trajectory space ($2^{50} \approx 10^{15}$ paths), the action-value network uses two hidden layers of 64 units each ($64 \times 64$ MLP with Layer Normalization). Following standard benchmark convention~\citep{osband2019behaviour}, an environment run is deemed solved when the agent achieves consistent policy convergence, operationalized as maintaining a rolling mean episode return $> 0.9$ over 20 consecutive episodes (relative to the optimal return $1.0 - 0.01 = 0.99$), corresponding to the point where cumulative regret permanently flatlines. DP-DQN consistently solves DeepSea-50 in $1{,}925 \pm 184$ episodes (Table~\ref{tab:empirical_summary}). Across grid sizes from $N=10$ to $N=50$, empirical sample complexity scales as $\mathcal{O}(N^{1.45})$, outperforming the empirical $\mathcal{O}(N^3)$ scaling of BootDQN-RP~\citep{osband2018randomized} and remaining well below the $\Omega(2^N)$ lower bound of dithering heuristics.

\paragraph{Computational and Memory Efficiency.}
A foundational advantage of DP-DQN is that it eliminates the severe computational overhead of multi-network ensembles. While BootDQN-RP ($K=20$) must instantiate, store, and train $K$ separate action-value networks alongside $K$ prior networks and target networks ($\mathcal{O}(Kd)$ parameters and optimizer buffers), DP-DQN operates on a \emph{single active neural network} ($\mathcal{O}(d)$ parameters), matching the parameter footprint of standard DQN. The non-parametric mechanism introduces two computational overheads relative to standard DQN: (i)~sampling Dirichlet process stick-breaking weights requires $\mathcal{O}(T)$ scalar operations once per episode for prior atoms and $\mathcal{O}(B)$ per warmstart step for empirical mini-batches, totaling an asymptotic stick-breaking complexity of only $\mathcal{O}(T + W \cdot B)$ scalar operations per episode ($\mathcal{O}(T + WB) \ll W(B+T)d$, negligible compared to network tensor arithmetic); and (ii)~evaluating the composite loss (Eq.~\eqref{eq:dp_bellman_loss}) across the $T$ prior transitions in addition to the empirical mini-batch of size $B$ scales gradient backpropagation by a factor of $1 + T/B$. Profiling on canonical DeepSea-20 on a standardized CPU core confirms this efficiency: DP-DQN incurs an amortized execution latency of $0.43\,$ms/step (comparable to standard DQN's $0.38\,$ms/step, and $2.4\times$ faster than BootDQN-RP's $1.05\,$ms/step). Combined with superior sample efficiency ($602 \pm 104$ vs.\ $5{,}210 \pm 604$ episodes), DP-DQN converges in only $5.16\,$s of total CPU time compared to $103.26\,$s for BootDQN-RP—achieving a $20.0\times$ wall-clock speedup while using $20\times$ less memory (Table~\ref{tab:computational_complexity_deepsea20}, Appendix~\ref{sec:computational-complexity-analysis}).

\paragraph{Generality Across Diverse Decision Tasks.}
To evaluate whether the exploratory advantages of DP-BNNs generalize beyond DeepSea, we benchmarked the framework across diverse decision-making tasks (Table~\ref{tab:empirical_summary} and Appendix~\ref{sec:empirical-suite}). Across deceptive sequential corridors ($N$-Chain; \citealt{strens2000bayesian}), environmental stochasticity and reward traps (RiverSwim-6; \citealt{strehl2008analysis}), continuous control under gravity constraints (MountainCar-v0; \citealt{sutton1998reinforcement}), and deep contextual bandits (Wheel Bandit; \citealt{riquelme2018deep}; Figure~\ref{fig:wheel_bandit_showdown}), DP-BNN posteriors consistently sustain directed exploration, matching or exceeding competitive multi-network baselines while operating with a single network. Complete experimental protocols and tabular comparisons are detailed in Appendix~\ref{sec:empirical-suite}.

\section{Conclusions and Future Work}
\label{sec:conclusions}

We introduced Dirichlet Process Bayesian Neural Networks (DP-BNNs), which quantify epistemic uncertainty by placing a nonparametric prior on the data-generating distribution rather than on network weights. We illustrated the application of DP-BNNs to deep reinforcement learning by developing DP-DQN for randomized value functions. By drawing an action-value hypothesis for each episode rollout, DP-DQN realizes temporally consistent Thompson sampling with a single neural network, resolving hard exploration challenges while avoiding the memory and computational overhead of multi-network ensembles.

Several promising research directions emerge from this paradigm. On the theoretical side, establishing formal finite-sample regret bounds for DP-DQN that explicitly capture the dependence on the DP prior hyperparameters—namely the concentration parameter $\alpha$, the base measure $F_0$, and the truncation level $T$—presents a vital avenue of inquiry to rigorously characterize how prior mass and truncation accuracy govern exploration efficiency and asymptotic convergence~\citep{osband2017posterior}. More broadly, designing structured base measures to incorporate domain-specific inductive biases presents an exciting future direction~\citep{klissarov2023deep, machado2023temporal}, which is notoriously difficult to achieve in weight-space Bayesian neural networks~\citep{wilson2020bayesian}. Finally, extending DP-BNN posteriors to continuous actor-critic architectures holds strong practical promise to eliminate the multi-critic ensembles commonly required to mitigate value overestimation or estimate uncertainty~\citep{haarnoja2018soft}, enabling principled, sample-efficient continuous control with a single neural network.

\subsubsection*{Acknowledgments}
The author is grateful to the Action Editor and the anonymous reviewers for their insightful and constructive reviews, which significantly improved the paper. The author warmly thanks Dr.~Igor Pr{\"u}nster, Dr.~Pierre Alquier, and Dr.~Odalric-Ambrym Maillard for encouraging comments. 

\subsubsection*{Reproducibility Statement}
The source code, environment wrappers, configuration files, reproduction scripts, and pedagogical Python notebooks are available at \url{https://github.com/sumilations/Dirichlet-Process-Bayesian-Neural-networks}. Experimental hyperparameters, architectures, and evaluation protocols are detailed in Appendix~\ref{sec:empirical-suite} (Table~\ref{tab:hyperparameters}), while mathematical derivations and algorithmic analyses are provided in Appendices~\ref{sec:stick-breaking-finite}--\ref{sec:stochastic-optimism-and-bdqn-failure}.

\bibliography{main}

@book{mackay2003information,
  title={Information theory, inference and learning algorithms},
  author={MacKay, David JC},
  year={2003},
  publisher={Cambridge university press}
}

@inproceedings{blundell2015weight,
  title={Weight uncertainty in neural network},
  author={Blundell, Charles and Cornebise, Julien and Kavukcuoglu, Koray and Wierstra, Daan},
  booktitle={International conference on machine learning},
  pages={1613--1622},
  year={2015},
  organization={PMLR}
}

@article{arbel2023primer,
  title={A Primer on {B}ayesian Neural Networks: Review and Debates},
  author={Arbel, Julyan and Pitas, Konstantinos and Vladimirova, Mariia and Fortuin, Vincent},
  journal={arXiv preprint arXiv:2309.16314},
  year={2023}
}

@inproceedings{gal2016dropout,
  title={Dropout as a {B}ayesian Approximation: Representing Model Uncertainty in Deep Learning},
  author={Gal, Yarin and Ghahramani, Zoubin},
  booktitle={international conference on machine learning},
  pages={1050--1059},
  year={2016},
  organization={PMLR}
}

@article{mnih2015human,
  title={Human-level control through deep reinforcement learning},
  author={Mnih, Volodymyr and Kavukcuoglu, Koray and Silver, David and Rusu, Andrei A and Veness, Joel and Bellemare, Marc G and Graves, Alex and Riedmiller, Martin and Fidjeland, Andreas K and Ostrovski, Georg and others},
  journal={nature},
  volume={518},
  number={7540},
  pages={529--533},
  year={2015},
  publisher={Nature Publishing Group}
}

@inproceedings{bellemare2017distributional,
  title={A distributional perspective on reinforcement learning},
  author={Bellemare, Marc G and Dabney, Will and Munos, R{\'e}mi},
  booktitle={International conference on machine learning},
  pages={449--458},
  year={2017},
  organization={PMLR}
}

@article{vashishtha2025leveraging,
    title={Leveraging priors on distribution functions for multi-arm bandits},
    author={Vashishtha, Sumit and Maillard, Odalric-Ambrym},
    journal={Reinforcement Learning Journal},
    volume={6},
    pages={1600--1623},
    year={2025}
}

@article{puterman1990markov,
  title={{M}arkov Decision Processes},
  author={Puterman, Martin L},
  journal={Handbooks in operations research and management science},
  volume={2},
  pages={331--434},
  year={1990},
  publisher={Elsevier}
}

@incollection{ghosal2010dirichlet,
  title={The {D}irichlet process, related priors and posterior asymptotics},
  author={Ghosal, Subhashis},
  booktitle={Bayesian Nonparametrics},
  editor={Hjort, Nils Lid and Holmes, Chris and M{\"u}ller, Peter and Walker, Stephen G.},
  pages={35--79},
  year={2010},
  publisher={Cambridge University Press}
}

@article{muliere1998approximating,
  title={Approximating Distributions of Random Functionals of {F}erguson--{D}irichlet Priors},
  author={Muliere, Pietro and Tardella, Luca},
  journal={Canadian Journal of Statistics},
  volume={26},
  number={2},
  pages={283--297},
  year={1998},
  publisher={Wiley Online Library}
}

@article{thompson1933likelihood,
  title={On the likelihood that one unknown probability exceeds another in view of the evidence of two samples},
  author={Thompson, William R},
  journal={Biometrika},
  volume={25},
  number={3-4},
  pages={285--294},
  year={1933},
  publisher={Oxford University Press}
}

@article{ishwaran2001gibbs,
  title={Gibbs sampling methods for stick-breaking priors},
  author={Ishwaran, Hemant and James, Lancelot F},
  journal={Journal of the American statistical Association},
  volume={96},
  number={453},
  pages={161--173},
  year={2001},
  publisher={Taylor \& Francis}
}

@article{gaffi2025random,
  title={Random probability measures with fixed mean distributions},
  author={Gaffi, Francesco and Lijoi, Antonio and Pr{\"u}nster, Igor},
  journal={The Annals of Applied Probability},
  volume={35},
  number={4},
  pages={2239--2261},
  year={2025},
  publisher={Institute of Mathematical Statistics}
}

@article{lijoi2009distributional,
  title={Distributional properties of means of random probability measures},
  author={Lijoi, Antonio and Pr{\"u}nster, Igor},
  journal={Statistics Surveys},
  volume={3},
  pages={47--95},
  year={2009},
  publisher={Amer. Statist. Assoc., Bernoulli Soc., Inst. Math. Statist., and Soc. Fran{\c{c}}aise Stat.}
}

@inproceedings{kveton2019garbage,
  title={Garbage In, Reward Out: Bootstrapping Exploration in Multi-Armed Bandits},
  author={Kveton, Branislav and Szepesvari, Csaba and Vaswani, Sharan and Wen, Zheng and Lattimore, Tor and Ghavamzadeh, Mohammad},
  booktitle={International Conference on Machine Learning},
  pages={3601--3610},
  year={2019},
  organization={PMLR}
}

@book{efron1994introduction,
  title={An introduction to the bootstrap},
  author={Efron, Bradley and Tibshirani, Robert J},
  year={1994},
  publisher={Chapman and Hall/CRC}
}

@inproceedings{kaufmann2012thompson,
  title={{T}hompson Sampling: An Asymptotically Optimal Finite-Time Analysis},
  author={Kaufmann, Emilie and Korda, Nathaniel and Munos, R{\'e}mi},
  booktitle={International Conference on Algorithmic Learning Theory (ALT)},
  pages={199--213},
  year={2012},
  organization={Springer}
}

@article{russo2018tutorial,
  title={A Tutorial on {T}hompson Sampling},
  author={Russo, Daniel J and Van Roy, Benjamin and Kazerouni, Abbas and Osband, Ian and Wen, Zheng and others},
  journal={Foundations and Trends{\textregistered} in Machine Learning},
  volume={11},
  number={1},
  pages={1--96},
  year={2018},
  publisher={Now Publishers, Inc.}
}

@article{osband2019deep,
  title={Deep exploration via randomized value functions},
  author={Osband, Ian and Van Roy, Benjamin and Russo, Daniel J and Wen, Zheng},
  journal={Journal of Machine Learning Research},
  volume={20},
  number={124},
  pages={1--62},
  year={2019}
}

@article{osband2016deep,
  title={Deep Exploration via Bootstrapped {DQN}},
  author={Osband, Ian and Blundell, Charles and Pritzel, Alexander and Van Roy, Benjamin},
  journal={Advances in neural information processing systems},
  volume={29},
  year={2016}
}

@article{roeder1990density,
  title={Density estimation with confidence sets exemplified by superclusters and voids in the galaxies},
  author={Roeder, Kathryn},
  journal={Journal of the American Statistical Association},
  volume={85},
  number={411},
  pages={617--624},
  year={1990},
  publisher={Taylor \& Francis}
}

@article{osband2013more,
  title={(More) efficient reinforcement learning via posterior sampling},
  author={Osband, Ian and Russo, Daniel and Van Roy, Benjamin},
  journal={Advances in Neural Information Processing Systems},
  volume={26},
  year={2013}
}

@article{ferguson1973bayesian,
  title={A {B}ayesian analysis of some nonparametric problems},
  author={Ferguson, Thomas S.},
  journal={The Annals of Statistics},
  volume={1},
  number={2},
  pages={209--230},
  year={1973},
  publisher={Institute of Mathematical Statistics}
}

@article{rubin1981bayesian,
  title={The {B}ayesian {B}ootstrap},
  author={Rubin, Donald B.},
  journal={The Annals of Statistics},
  volume={9},
  number={1},
  pages={130--134},
  year={1981},
  publisher={Institute of Mathematical Statistics}
}

@article{sethuraman1994constructive,
  title={A constructive definition of {D}irichlet priors},
  author={Sethuraman, Jayaram},
  journal={Statistica Sinica},
  volume={4},
  number={2},
  pages={639--650},
  year={1994}
}

@book{ghosal2017fundamentals,
  title={Fundamentals of Nonparametric {B}ayesian Inference},
  author={Ghosal, Subhashis and van der Vaart, Aad W},
  volume={44},
  year={2017},
  publisher={Cambridge University Press}
}

@inproceedings{osband2017posterior,
  title={Why is posterior sampling better than optimism for reinforcement learning?},
  author={Osband, Ian and Van Roy, Benjamin},
  booktitle={International conference on machine learning},
  pages={2701--2710},
  year={2017},
  organization={PMLR}
}

@article{auer2008near,
  title={Near-optimal regret bounds for reinforcement learning},
  author={Auer, Peter and Jaksch, Thomas and Ortner, Ronald},
  journal={Advances in neural information processing systems},
  volume={21},
  year={2008}
}

@article{osband2018randomized,
  title={Randomized prior functions for deep reinforcement learning},
  author={Osband, Ian and Aslanides, John and Cassirer, Albin},
  journal={Advances in neural information processing systems},
  volume={31},
  year={2018}
}

@phdthesis{maillard2019mathematics,
  title={Mathematics of statistical sequential decision making},
  author={Maillard, Odalric-Ambrym},
  type={{HDR} dissertation},
  year={2019},
  school={Universit{\'e} de Lille}
}

@article{brafman2002r,
  title={R-max-a general polynomial time algorithm for near-optimal reinforcement learning},
  author={Brafman, Ronen I and Tennenholtz, Moshe},
  journal={Journal of Machine Learning Research},
  volume={3},
  number={Oct},
  pages={213--231},
  year={2002}
}

@article{lakshminarayanan2017simple,
  title={Simple and scalable predictive uncertainty estimation using deep ensembles},
  author={Lakshminarayanan, Balaji and Pritzel, Alexander and Blundell, Charles},
  journal={Advances in neural information processing systems},
  volume={30},
  year={2017}
}

@book{sutton1998reinforcement,
  title={Reinforcement learning: An introduction},
  author={Sutton, Richard S and Barto, Andrew G },
  volume={1},
  number={1},
  year={1998},
  publisher={MIT press Cambridge}
}

@article{riquelme2018deep,
  title={Deep {B}ayesian Bandits Showdown: An Empirical Comparison of {B}ayesian Deep Networks for {T}hompson Sampling},
  author={Riquelme, Carlos and Tucker, George and Snoek, Jasper},
  journal={arXiv preprint arXiv:1802.09127},
  year={2018}
}

@inproceedings{welling2011bayesian,
  title={{B}ayesian Learning via Stochastic Gradient {L}angevin Dynamics},
  author={Welling, Max and Teh, Yee W},
  booktitle={Proceedings of the 28th international conference on machine learning (ICML-11)},
  pages={681--688},
  year={2011}
}

@phdthesis{vashishtha2026bayesian,
  title={{B}ayesian Reinforcement Learning with {D}irichlet Processes},
  author={Vashishtha, Sumit},
  year={2026},
  school={Inria \& Universit{\'e} de Lille}
}

@article{osband2019behaviour,
  title={Behaviour suite for reinforcement learning},
  author={Osband, Ian and Doron, Yotam and Hessel, Matteo and Aslanides, John and Sezener, Eren and Saraiva, Andre and McKinney, Katrina and Lattimore, Tor and Szepesvari, Csaba and Singh, Satinder and others},
  journal={arXiv preprint arXiv:1908.03568},
  year={2019}
}

@inproceedings{agrawal2012analysis,
  title={Analysis of {T}hompson sampling for the multi-armed bandit problem},
  author={Agrawal, Shipra and Goyal, Navin},
  booktitle={Conference on Learning Theory (COLT)},
  pages={39--1},
  year={2012},
  organization={JMLR Workshop and Conference Proceedings}
}

@inproceedings{agrawal2017optimistic,
  title={Optimistic posterior sampling for reinforcement learning: worst-case regret bounds},
  author={Agrawal, Shipra and Jia, Randy},
  booktitle={Advances in Neural Information Processing Systems (NeurIPS)},
  volume={30},
  year={2017}
}

@inproceedings{azizzadenesheli2018efficient,
  title={Efficient Exploration through {B}ayesian Deep {Q}-Networks},
  author={Azizzadenesheli, Kamyar and Brunskill, Emma and Anandkumar, Animashree},
  booktitle={2018 Information Theory and Applications Workshop (ITA)},
  pages={1--9},
  year={2018},
  organization={IEEE}
}

@article{houlsby2011bayesian,
  title={{B}ayesian Active Learning for Classification and Preference Learning},
  author={Houlsby, Neil and Husz{\'a}r, Ferenc and Ghahramani, Zoubin and Lengyel, M{\'a}t{\'e}},
  journal={arXiv preprint arXiv:1112.5745},
  year={2011}
}

@article{strehl2008analysis,
  title={An Analysis of Model-Based Interval Estimation for {M}arkov Decision Processes},
  author={Strehl, Alexander L. and Littman, Michael L.},
  journal={Journal of Computer and System Sciences},
  volume={74},
  number={8},
  pages={1309--1331},
  year={2008},
  publisher={Elsevier}
}

@inproceedings{strens2000bayesian,
  title={A {B}ayesian Approach to Reinforcement Learning},
  author={Strens, Malcolm},
  booktitle={Proceedings of the Seventeenth International Conference on Machine Learning (ICML)},
  pages={943--950},
  year={2000}
}

@article{lillicrap2015continuous,
  title={Continuous control with deep reinforcement learning},
  author={Lillicrap, Timothy P and Hunt, Jonathan J and Pritzel, Alexander and Heess, Nicolas and Erez, Tom and Tassa, Yuval and Silver, David and Wierstra, Daan},
  journal={arXiv preprint arXiv:1509.02971},
  year={2015}
}

@article{ba2016layer,
  title={Layer normalization},
  author={Ba, Jimmy Lei and Kiros, Jamie Ryan and Hinton, Geoffrey E},
  journal={arXiv preprint arXiv:1607.06450},
  year={2016}
}

@inproceedings{klissarov2023deep,
  title     = {Deep Laplacian-based Options for Temporally-Extended Exploration},
  author    = {Klissarov, Martin and Machado, Marlos C.},
  booktitle = {International Conference on Machine Learning (ICML)},
  pages     = {17067--17088},
  year      = {2023},
  organization = {PMLR}
}

@article{machado2023temporal,
  author  = {Machado, Marlos C. and Barreto, Andr{\'{e}} and Precup, Doina and Bowling, Michael},
  title   = {Temporal Abstraction in Reinforcement Learning with the Successor Representation},
  journal = {Journal of Machine Learning Research},
  year    = {2023},
  volume  = {24},
  number  = {80},
  pages   = {1--69}
}

@inproceedings{foong2019inbetween,
  title={``{I}n-between'' uncertainty in {B}ayesian neural networks},
  author={Foong, Andrew YK and Li, Yingzhen and Hern{\'a}ndez-Lobato, Jos{\'e} Miguel and Turner, Richard E},
  booktitle={ICML Workshop on Uncertainty and Robustness in Deep Learning},
  year={2019}
}

@article{ernst2005tree,
  title={Tree-based batch mode reinforcement learning},
  author={Ernst, Damien and Geurts, Pierre and Wehenkel, Louis},
  journal={Journal of Machine Learning Research},
  volume={6},
  number={Apr},
  pages={503--556},
  year={2005}
}

@inproceedings{riedmiller2005neural,
  title={Neural fitted {Q} iteration--first experiences with a data efficient neural reinforcement learning method},
  author={Riedmiller, Martin},
  booktitle={European Conference on Machine Learning},
  pages={317--328},
  year={2005},
  organization={Springer}
}

@inproceedings{dearden1998bayesian,
  title={Bayesian {Q}-learning},
  author={Dearden, Richard and Friedman, Nir and Andre, David},
  booktitle={AAAI/IAAI},
  pages={761--768},
  year={1998}
}

@inproceedings{dwaracherla2020hypermodels,
  title={Hypermodels for Exploration},
  author={Dwaracherla, Vikranth and Asadi, Kavosh and Osband, Ian},
  booktitle={International Conference on Learning Representations},
  year={2020}
}

@inproceedings{fortunato2018noisy,
  title={Noisy Networks for Exploration},
  author={Fortunato, Meire and Azar, Mohammad Gheshlaghi and Piot, Bilal and Menick, Jacob and Osband, Ian and Graves, Alex and Mnih, Vlad and Muldal, Alistair and Degris, Thomas and Legg, Shane and Blundell, Charles and Leike, Jan},
  booktitle={International Conference on Learning Representations},
  year={2018}
}

@inproceedings{haarnoja2018soft,
  title={Soft actor-critic: Off-policy maximum entropy deep reinforcement learning with a stochastic actor},
  author={Haarnoja, Tuomas and Zhou, Aurick and Abbeel, Pieter and Levine, Sergey},
  booktitle={International Conference on Machine Learning (ICML)},
  pages={1861--1870},
  year={2018}
}

@inproceedings{wilson2020bayesian,
  title     = {{B}ayesian Deep Learning and a Probabilistic Perspective of Generalization},
  author    = {Wilson, Andrew Gordon and Izmailov, Pavel},
  booktitle = {Advances in Neural Information Processing Systems (NeurIPS)},
  volume    = {33},
  pages     = {4697--4708},
  year      = {2020}
}

@inproceedings{oren2025epistemic,
  title     = {Epistemic {M}onte {C}arlo Tree Search},
  author    = {Oren, Yaniv and Vadocz, Viliam and Spaan, Matthijs T. J. and Boehmer, Wendelin},
  booktitle = {International Conference on Learning Representations (ICLR)},
  year      = {2025}
}

@book{castillo2024bayesian,
  title     = {{B}ayesian Nonparametric Statistics: {\'E}cole d'{\'E}t{\'e} de Probabilit{\'e}s de Saint-Flour LI--2023},
  author    = {Castillo, Isma{\"e}l},
  series    = {Lecture Notes in Mathematics},
  volume    = {2358},
  publisher = {Springer},
  year      = {2024}
}

@inproceedings{van2016deep,
  title     = {Deep Reinforcement Learning with Double {Q}-learning},
  author    = {van Hasselt, Hado and Guez, Arthur and Silver, David},
  booktitle = {AAAI Conference on Artificial Intelligence},
  volume    = {30},
  number    = {1},
  pages     = {2094--2100},
  year      = {2016}
}

@article{ghavamzadeh2015bayesian,
  title     = {{B}ayesian Reinforcement Learning: A Survey},
  author    = {Ghavamzadeh, Mohammad and Mannor, Shie and Pineau, Joelle and Tamar, Aviv},
  journal   = {Foundations and Trends{\textregistered} in Machine Learning},
  volume    = {8},
  number    = {5-6},
  pages     = {359--483},
  year      = {2015}
}

@inproceedings{durand2018streaming,
  title     = {Streaming Kernel Regression with Provably Adaptive Mean, Variance, and Regularization},
  author    = {Durand, Audrey and Maillard, Odalric-Ambrym and Pineau, Joelle},
  booktitle = {International Conference on Artificial Intelligence and Statistics (AISTATS)},
  pages     = {1849--1857},
  year      = {2018}
}
\bibliographystyle{tmlr}

\clearpage
\appendix

\begin{center}
    \hrule height 1.5pt
    \vskip 0.15in
    {\LARGE \bf Supplementary Material: Technical Appendix} \\[0.4em]
    {\large \textbf{Bayesian Neural Networks via Dirichlet Process Priors on the Data-Generating Distribution for Randomized Value Functions in Deep Reinforcement Learning}}
    \vskip 0.15in
    \hrule height 1pt
\end{center}
\vspace{1.5em}

\paragraph{Organization of the Appendix.} The supplementary appendix is organized as follows:
\begin{itemize}[leftmargin=*,noitemsep]
    \item \textbf{Appendix~\ref{sec:stick-breaking-finite}:} Truncated stick-breaking representation, truncation bounds, the 99.9\% residual mass rule, and galaxy velocity density estimation.
    \item \textbf{Appendix~\ref{sec:DPPSHyperparameters}:} Base measure topological support conditions, 2D annular manifold benchmark formulation, and closed-form BALD derivations.
    \item \textbf{Appendix~\ref{sec:stochastic-warmstart-theory-appendix}:} Dynamic stochastic mini-batch warmstart mechanics, prevention of prior attenuation, and Bellman credit propagation depth.
    \item \textbf{Appendix~\ref{sec:stochastic-optimism-and-bdqn-failure}:} Stochastic optimism and deep exploration analysis, contrasting DP-BNNs against randomized prior ensembles (BootDQN-RP) and neural-linear approximations (BDQN), with empirical base measure ablations.
    \item \textbf{Appendix~\ref{sec:empirical-suite}:} Consolidated empirical benchmark suite (Table~\ref{tab:empirical_summary}) and hyperparameter protocols (Table~\ref{tab:hyperparameters}) covering DeepSea, RiverSwim-6, $N$-Chain, and Wheel Bandit.
    \item \textbf{Appendix~\ref{sec:mountaincar-continuous}:} Continuous state-space RL exploration on MountainCar-v0 with a model-free Brownian diffusive base measure.
    \item \textbf{Appendix~\ref{sec:computational-complexity-analysis}:} Computational and memory complexity comparison across value-based deep RL paradigms (Table~\ref{tab:computational_complexity_theoretical}).
\end{itemize}

\section{Truncated Stick-Breaking Representation and Hyperparameter Selection}
\label{sec:stick-breaking-finite}
\label{sec:truncation-999-rule}

The finite stick-breaking representation of DP priors (Equations~\eqref{def:SB-DP}--\eqref{def:SB-DP-3}) and the posterior random measure construction~\citep{vashishtha2025leveraging} provide an exact, constructive generative mechanism for sampling random probability measures~\citep{sethuraman1994constructive, ishwaran2001gibbs}. Rather than relying on heuristic or arbitrary cutoff levels, we determine the stick-breaking truncation level $T$ (the number of prior pseudo-observations) for the prior measure $\mathrm{DP}(\alpha, F_0)$ using a principled 99.9\% expected residual probability mass rule.

Under Sethuraman's stick-breaking construction, prior probability weights are generated via $\pi_k = V_k \prod_{l < k} (1 - V_l)$ with $V_l \stackrel{\text{i.i.d.}}{\sim} \mathrm{Beta}(1, \alpha)$. When the series is truncated at $T$ atoms, the residual probability mass unallocated to the truncated support is $R_T = \sum_{k = T + 1}^\infty \pi_k = \prod_{k=1}^T (1 - V_k)$. Since $\mathbb{E}[1 - V_k] = \frac{\alpha}{\alpha + 1}$, mutual independence of $\{V_k\}_{k=1}^T$ yields the exact closed-form expected residual mass:
\begin{equation}
\mathbb{E}[R_T] = \prod_{k=1}^T \mathbb{E}[1 - V_k] = \left( \frac{\alpha}{\alpha + 1} \right)^T.
\end{equation}
To ensure that the finite truncation captures at least 99.9\% of the total prior probability mass ($\mathbb{E}[R_T] \le 10^{-3}$), we require:
\begin{equation}
\mathbb{E}[R_T] = \left( \frac{\alpha}{\alpha + 1} \right)^T \le 10^{-3} \implies T \ge \frac{\ln(1000)}{\ln\left( 1 + \frac{1}{\alpha} \right)} \approx \frac{6.9078}{\ln\left( 1 + \frac{1}{\alpha} \right)}.
\label{eq:999-truncation-bound}
\end{equation}

Equation~\eqref{eq:999-truncation-bound} provides an explicit, quantitative guideline for configuring $T$ across our benchmark suite:

\paragraph{Reinforcement Learning Tasks ($\alpha = 3.0$).} In DeepSea and $N$-Chain, evaluating Equation~\eqref{eq:999-truncation-bound} gives $T \ge \frac{6.9078}{\ln(4/3)} \approx 24.01$. Hence, any truncation level $T \ge 25$ guarantees $>99.9\%$ prior mass retention. In our standard implementation, setting $T = 50$ captures over 99.9999\% of the prior mass ($\mathbb{E}[R_T] = (0.75)^{50} \approx 5.66 \times 10^{-7} < 10^{-6}$), completely eliminating truncation distortion while incurring negligible computational overhead.

\paragraph{Unit Concentration Baseline ($\alpha = 1.0$).} Under standard uninformative Dirichlet priors ($\alpha = 1.0$), Equation~\eqref{eq:999-truncation-bound} yields $T \ge \frac{6.9078}{\ln(2)} \approx 9.96$, showing that just 10 atoms suffice to capture 99.9\% of the prior mass.

\paragraph{Dispersed Regression Benchmark ($\alpha = 10.0$).} In the 2D annular manifold benchmark (Appendix~\ref{sec:bald-details}), where high concentration $\alpha = 10.0$ is deliberately selected to produce dispersed functional hypotheses in the cavity void, Equation~\eqref{eq:999-truncation-bound} yields $T \ge \frac{6.9078}{\ln(1.10)} \approx 72.5$. Setting $T = 50$ captures over 99.1\% of the total prior mass ($\mathbb{E}[R_T] = (10/11)^{50} \approx 0.0085$).

\begin{figure}[h]
\centering
\includegraphics[width=0.6\linewidth]{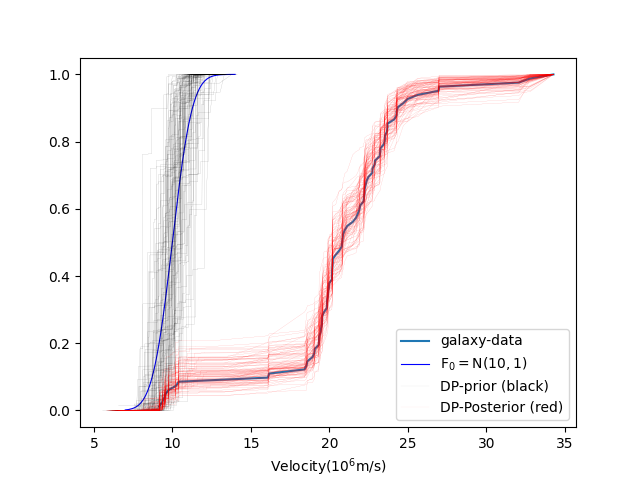}
\caption{DP prior and posterior density estimates on the galaxy velocity dataset.}
\label{fig:galaxy-dataset-DPprior/posterior}
\end{figure}

\paragraph{Empirical Density Estimation on the Galaxy Velocity Dataset.}
To validate our stick-breaking posterior sampler empirically, we evaluate density estimation on the canonical galaxy velocity benchmark~\citep{roeder1990density}. The measurements are velocities at which galaxies in the Corona-Borealis region move away from our galaxy. When galaxies are clustered, the velocity density is multimodal, as evidenced in the empirical CDF in Figure~\ref{fig:galaxy-dataset-DPprior/posterior} where mass concentrates in distinct interior modes. Starting with a $\mathrm{DP}(\alpha=2, F_{0}=\mathcal{N}(10, 1))$ prior, the spread of distributions sampled from the DP posterior provides a well-calibrated confidence band of the density estimated through the DP.

\section{Prior Base Measure Design and Epistemic Uncertainty Formulation}
\label{sec:DPPSHyperparameters}
\label{sec:dp-dqn-base-choice}

In this section, we formalize the topological support conditions required for the DP prior base measure $F_0$ and detail its concrete realization for epistemic uncertainty quantification in spatial regression.

\subsection{Base Measure Support Condition}
\label{sec:support-condition-dp}

In the initial exploration phase of learning, whether in supervised regression or reinforcement learning, observations are sparse or absent across entire regions of the input space. Value predictions and exploratory hypotheses are therefore driven almost entirely by the synthetic pseudo-observations generated through the DP prior base measure $F_0$. Because early exploratory actions have delayed, irreversible consequences in deep MDPs, the structure and domain support of $F_0$ are critical.

A natural and general formulation is to specify a factorized product base measure over transition tuples $z = (s, a, r, s', d)$:
\begin{equation}
F_0(s, a, r, s', d) = F_{0s}(s) \times F_{0a}(a) \times F_{0r}(r) \times F_{0s'}(s') \times F_{0d}(d).
\end{equation}
The specification of state, action, and terminal components $F_{0s}$, $F_{0a}$, and $F_{0d}$ must strictly respect the physical domain and support of the underlying Markov Decision Process (with $d=0$ for non-terminal continuing transitions). This requirement is mathematically dictated by the topological support properties of DPs:

\begin{lem}[Support of Dirichlet Processes~\citep{ghosal2010dirichlet}] \label{thm:Support-of-DPs} 
In the weak topology, the topological support of the DP $\mathrm{DP}(\alpha, F_0)$ is the set of all probability measures $P^\star$ whose supports are contained within the support of the base measure:
\begin{equation}
\mathrm{supp}(\mathrm{DP}(\alpha, F_0)) = \left\{ P^\star \in \mathcal{M}_1(\mathcal{X}) : \mathrm{supp}(P^\star) \subseteq \mathrm{supp}(F_0) \right\}.
\end{equation}
\end{lem}

Lemma~\ref{thm:Support-of-DPs} establishes that the DP posterior can place non-zero probability mass in an exploratory region if and only if that region is contained within the support of the base measure $F_0$. In practice, the factorized components are configured as follows:

\paragraph{Action Space Prior ($F_{0a}$).} For discrete action spaces with $|\mathcal{A}|$ actions, $F_{0a}$ is chosen as a uniform categorical distribution over $\{1, \dots, |\mathcal{A}|\}$. For continuous action spaces, $F_{0a}$ is selected to span the admissible action bounds (e.g., $\mathrm{Uniform}[a_{\min}, a_{\max}]$).

\paragraph{State Space and Transition Priors ($F_{0s}, F_{0s'}$).} Chosen to encapsulate the admissible state space $\mathcal{S}$ (e.g., uniform over coordinate bounds or standardized feature bounding boxes).

\paragraph{Reward Prior ($F_{0r}$).} Crucially, unlike supervised regression where observations are static, in reinforcement learning data collection is actively dictated by the agent's policy: a pessimistic reward prior discourages exploration, preventing the agent from ever gathering experience in unvisited states. Therefore, $F_{0r}$ is designed not to replicate environment step penalties (which are learned exclusively through the empirical replay stream $\mathbb{P}_n$), but to induce \emph{stochastic optimism}~\citep{osband2017posterior}: either via the uniform prior $F_{0r} = \mathcal{U}(0, R_{\max})$ on non-negative returns, or via zero-mean Gaussians $\mathcal{N}(0, \sigma_0^2)$ where extreme-value order statistics across $T$ prior atoms ($\mathbb{E}[\max_{j=1}^T \tilde{r}_j] \approx \sigma_0 \sqrt{2 \ln T} > 0$) naturally stimulate directed exploration without requiring knowledge of $R_{\max}$ or goal locations (Appendix~\ref{sec:stochastic-optimism-and-bdqn-failure}).

We now demonstrate how these foundational principles—the 99.9\% mass truncation rule and the topological support condition—are concretely instantiated in static spatial regression. Appendix~\ref{sec:bald-details} details the geometry, factorized base measure design, and information-theoretic BALD epistemic uncertainty derivations for the 2D annular benchmark introduced in Section~\ref{sec:DP-BNNs}.

\subsection{Prior Base Measure and BALD Metric for the 2D Regression Benchmark}
\label{sec:bald-details}

In this section, we provide the full mathematical formulation, prior base measure design, and hyperparameter specifications for the 2D epistemic uncertainty benchmark presented in Section~\ref{sec:DP-BNNs} and Figure~\ref{fig:2D-regression-uncertainty}.

\paragraph{Annular Manifold Geometry.} Training data points $\{x_i, y_i\}_{i=1}^n$ ($n = 300$) are sampled uniformly on an annulus with 2D spatial coordinates $x = (x_1, x_2) \in [-2.5, 2.5]^2$ and radial distance $r = \|x\|_2 \in [1.3, 2.45]$. The scalar targets $y \in [-1, 1]$ are generated via $y = \cos(1.5\pi r) + \epsilon$ with additive Gaussian observation noise $\epsilon \sim \mathcal{N}(0, \sigma^2_{\rm aleatoric})$ where $\sigma_{\rm aleatoric} = 0.30$. The central circular region $r \le 1.0$ is completely void of training data (Figure~\ref{fig:2D-regression-uncertainty}A). A well-calibrated epistemic uncertainty estimator must detect this out-of-distribution cavity by predicting high uncertainty in the void $\mathcal{V} = \{x \in [-2.5, 2.5]^2 : \|x\|_2 \le 1.0\}$ and low uncertainty on the data support $\mathcal{D} = \{x \in [-2.5, 2.5]^2 : 1.3 \le \|x\|_2 \le 2.45\}$.

\paragraph{Bayesian Active Learning by Disagreement (BALD).} Epistemic uncertainty is quantified information-theoretically using BALD~\citep{houlsby2011bayesian}, which measures the mutual information between the model output $y$ and the model parameters $\theta$ conditioned on the dataset $\mathcal{D}$:
\begin{equation}\label{eq:bald-definition}
\text{BALD}(x) \triangleq I(y, \theta \mid x, \mathcal{D}) = H(y \mid x, \mathcal{D}) - \mathbb{E}_{\theta \sim p(\theta \mid \mathcal{D})}[H(y \mid x, \theta)].
\end{equation}
Intuitively, BALD isolates the information that learning the true model parameters $\theta$ would provide about the prediction at test input $x$, cleanly separating epistemic uncertainty from inherent aleatoric observation noise $H(y \mid x, \theta)$. Under a Gaussian predictive distribution $\mathcal{N}(f_\theta(x), \sigma^2_{\rm aleatoric})$, the total predictive entropy across an ensemble of $J$ posterior models $\{\theta^{(j)}\}_{j=1}^J$ is given by:
\begin{equation}
H(y \mid x, \mathcal{D}) \approx \frac{1}{2} \ln \left( 2\pi e \left( \sigma^2_{\rm epistemic}(x) + \sigma^2_{\rm aleatoric} \right) \right),
\end{equation}
where $\sigma^2_{\rm epistemic}(x) = \frac{1}{J}\sum_{j=1}^J (f_{\theta^{(j)}}(x) - \bar{f}(x))^2$ is the ensemble variance, with mean prediction $\bar{f}(x) = \frac{1}{J}\sum_{j=1}^J f_{\theta^{(j)}}(x)$. The expected observation noise entropy is:
\begin{equation}
\mathbb{E}_{\theta}[H(y \mid x, \theta)] = \frac{1}{2} \ln (2\pi e \sigma^2_{\rm aleatoric}).
\end{equation}
Subtracting the two terms yields the closed-form expression for BALD in nats:
\begin{equation}\label{eq:bald-gaussian}
\text{BALD}(x) = \frac{1}{2} \ln \left( 1 + \frac{\sigma^2_{\rm epistemic}(x)}{\sigma^2_{\rm aleatoric}} \right).
\end{equation}
This metric evaluates strictly to zero when all posterior models agree ($\sigma^2_{\rm epistemic} = 0$), and scales logarithmically as posterior disagreement increases.

\paragraph{Architecture and Training Hyperparameters.}
All evaluated neural network models share an identical multi-layer perceptron architecture comprising 3 hidden layers with 64 units per layer and ReLU activations, without Layer Normalization, mapping $\mathbb{R}^2 \to \mathbb{R}$. Training is conducted using the Adam optimizer with a learning rate of $10^{-3}$ and batch size of 64 for 200 epochs across all models. For DP-BNN, posterior measure draws $P_j \sim \mathrm{DP}(\alpha_n, \bar{F}_n)$ are drawn with $J=10$ independent posterior samples.

\paragraph{Base Measure Construction and Prior Rationale.}
In accordance with Lemma~\ref{thm:Support-of-DPs}, the support of the DP prior must encapsulate the entire domain of interest to allow posterior mass in regions lacking empirical data. We specify a factorized product base measure $F_0(x, y) = F_{0x}(x) \times F_{0y}(y)$ consisting of: (i)~a continuous spatial input prior $F_{0x} = \mathcal{U}([-2.5, 2.5]^2)$, ensuring uniform prior coverage across both the observed annulus and the unobserved cavity void, and (ii)~a zero-mean Gaussian target prior $F_{0y} = \mathcal{N}(0, \sigma_0^2)$ with dispersed target variance ($\sigma_0 = 1.5 \gg \sigma_{\rm aleatoric} = 0.30$). We truncate the prior component at $T = 50$ synthetic pseudo-observations with concentration parameter $\alpha = 10.0$. Because the prior atoms sample diverse target values in the void with non-trivial Dirichlet weights, each posterior measure draw $P_j$ pulls the network towards a distinct hypothesis in the void, naturally producing the sharp epistemic bubble observed in Figure~\ref{fig:2D-regression-uncertainty}.

\section{Dynamic Stochastic Mini-Batch Warmstart and Algorithmic Dynamics}
\label{sec:stochastic-warmstart-theory-appendix}
\label{sec:dirichlet-generalization-appendix}

In this section, we provide a comprehensive analysis of the stochastic mini-batch warmstart mechanism introduced in Section~\ref{sec:DP-BNN-randomized-value-functions} and Algorithm~\ref{alg: deep-rl-dp}, detailing why conditioning directly on the full replay buffer fails in reinforcement learning and how dynamic stick-breaking updates resolve this fundamental challenge.

\subsection{Pathologies of Full-Buffer Conditioning and Static Freezing}
In classical Bayesian nonparametric regression, one is given a static dataset $\mathcal{D} = \{z_i\}_{i=1}^n$ and conditions the DP prior $\text{DP}(\alpha, F_0)$ directly on all $n$ observations, yielding the posterior:
\begin{equation}
P \mid \mathcal{D} \sim \text{DP}\left(\alpha + n, \; \frac{n}{\alpha + n} \mathbb{P}_n + \frac{\alpha}{\alpha + n} F_0\right),
\end{equation}
where $\mathbb{P}_n = \frac{1}{n} \sum_{i=1}^n \delta_{z_i}$ is the empirical distribution of observations. In reinforcement learning, however, the replay buffer $\mathcal{B}$ fundamentally violates the statistical assumptions of supervised learning:

\paragraph{Data Redundancy and Non-Uniformity.} Early RL rollouts are overwhelmingly dominated by repetitive transitions in absorbing, suboptimal regions of the state space (e.g., falling into bottom-left corners in DeepSea or cycling through uninformative states). The replay buffer rapidly accumulates tens of thousands of near-identical transitions ($M = |\mathcal{B}| \sim 10^5 - 10^6$).

\paragraph{Prior Swamping and Epistemic Starvation.} In standard full-buffer DP conditioning, the relative weight assigned to the prior base measure scales as:
\begin{equation}
\omega_{\text{prior}} = \frac{\alpha}{\alpha + M} \xrightarrow{M \gg \alpha} 0.
\end{equation}
While asymptotic concentration ($\omega_{\text{prior}} \to 0$) is desirable in stationary supervised learning, in exploration-sensitive RL it causes \textit{epistemic starvation}: the prior base measure $F_0$, which is specifically designed to inject optimism into unexplored states, is overwhelmed by the sheer volume of redundant failure data. As a consequence, the agent ceases to explore and prematurely collapses into pessimistic, greedy exploitation of suboptimal trajectories.

\paragraph{Computational and Memory Bottlenecks.} Constructing truncated stick-breaking weights $p_1, \dots, p_M$ and performing backpropagation through a deep neural network over $M \sim 10^5$ samples at every single episode imposes an intolerable computational and memory burden.

Conversely, sampling a single mini-batch $\mathcal{D} \subset \mathcal{B}$ of size $B \ll M$ and freezing it across all gradient steps creates severe empirical overfitting: the value function optimizes exclusively against an isolated empirical slice, ignoring the rest of the buffer and failing to propagate Bellman credit across multi-step trajectory chains.

\subsection{Resolution via Dynamic Stochastic Mini-Batch Warmstart}
Algorithm~\ref{alg: deep-rl-dp} overcomes the twin pathologies of computational overhead and prior swamping by performing stochastic optimization over dynamic, freshly sampled stick-breaking measures at episode boundaries.

Under the DP posterior, each sampled random measure decomposes into an empirical component over observed replay transitions and a prior component drawn from $F_0$. Rather than performing computationally prohibitive full-buffer optimization ($|\mathcal{B}| \sim 10^5$), the $W = \lceil H / 2 \rceil$ warmstart steps use mini-batch stochastic gradient descent to approximate the empirical component, while modulating prior salience online via Bellman temporal-difference surprise:

\paragraph{Episodic Prior Generation.} At the beginning of episode $k$, the agent samples synthetic prior transitions $\tilde{\mathcal{D}}_0 = \{(\tilde{s}_j, \tilde{a}_j, \tilde{r}_j, \tilde{s}'_j, \tilde{d}_j)\}_{j=1}^T \stackrel{\text{iid}}{\sim} F_0$ and computes prior stick weights $q_j^{(0)} = V'_j \prod_{l < j} (1 - V'_l)$ with $V'_j \sim \mathrm{Beta}(1, \alpha)$ for $j=1, \dots, T$. Because the warmstart phase is an iterative procedure approximating a single posterior draw, sampling the prior pseudo-atoms and stick weights once per episode instantiates a coherent, frozen Thompson sampling value hypothesis for the upcoming rollout.

\paragraph{Information Calibration via Bellman TD Surprise.} In sparse-reward exploration, unguided initial rollouts inevitably accumulate thousands of redundant, unrewarded transitions. Standard Bayesian updating scaling with the raw buffer size ($n = |\mathcal{B}| \sim 10^5$) would cause the prior weight to collapse prematurely ($\omega_{\text{prior}} \to 0$). From an information-theoretic standpoint, however, repeating predictable failure trajectories yields zero novel information about the environment. To prevent prior swamping, DP-DQN tracks an effective statistical sample count $N_{\text{stat}}$ calibrated online by the Bellman TD surprise:
\begin{equation}
\label{eq:appendix_n_stat}
N_{\text{stat}} \leftarrow N_{\text{stat}} + \frac{1}{B} \sum_{i=1}^B \min\left(1, \; \frac{|\delta_i|}{\sigma_{\text{TD}}}\right), \quad \text{where} \quad \delta_i = y_i - Q(s_i, a_i; \theta),
\end{equation}
and $\sigma_{\text{TD}}$ denotes the running standard deviation of Bellman errors. Transitions with near-zero Bellman error ($|\delta_i| \approx 0$) represent well-modeled dynamics that contribute negligible information, incrementing $N_{\text{stat}}$ by $\approx 0$. Only when the agent encounters statistically novel transitions ($|\delta_i| \ge \sigma_{\text{TD}}$) does $N_{\text{stat}}$ increment toward $1.0$.

\paragraph{Composite Stochastic Warmstart Updates.} Across $w = 1, \dots, W$ warmstart steps, the online network is optimized against the episodic hypothesis using dynamic empirical mini-batches. At each step $w$, a fresh empirical mini-batch $\mathcal{D}_w = \{(s_i, a_i, r_i, s'_i, d_i)\}_{i=1}^B \sim \mathcal{B}$ is sampled uniformly at random from the replay buffer. The total prior probability mass assigned to the synthetic base transitions is sampled as:
\begin{equation}
\label{eq:appendix_w_prior}
W_{\text{prior}} \sim \mathrm{Beta}(\alpha + 1, N_{\text{stat}}),
\end{equation}
yielding expected prior weight $\mathbb{E}[W_{\text{prior}}] = \frac{\alpha + 1}{\alpha + 1 + N_{\text{stat}}}$. The remaining empirical probability mass $(1 - W_{\text{prior}})$ is partitioned across the $B$ mini-batch transitions via the multi-step stick-breaking recursion (Eq.~\eqref{eq:multi-step-posterior}):
\begin{equation}
\label{eq:appendix_w_emp}
w_i^{\text{emp}} = (1 - W_{\text{prior}}) \frac{\tilde{w}_i}{\sum_{k=1}^B \tilde{w}_k}, \quad \text{where} \quad \tilde{w}_i = V_i \prod_{k=i+1}^B (1 - V_k), \quad V_i \sim \mathrm{Beta}(1, \alpha + i),
\end{equation}
with $\tilde{w}_B = V_B$, forming the discrete composite stick-breaking measure:
\begin{equation}
\label{eq:appendix_p_w}
P_w = \sum_{i=1}^B w_i^{\text{emp}} \delta_{z_i} + W_{\text{prior}} \sum_{j=1}^T q_j^{(0)} \delta_{\tilde{z}_j}.
\end{equation}
Minimizing the composite DP-Bellman loss $\mathcal{L}_{\text{DP}}(\theta)$ (Eq.~\eqref{eq:dp_bellman_loss}) via stochastic gradient descent updates online parameters $\theta$, while $N_{\text{stat}}$ is calibrated online via Eq.~\eqref{eq:appendix_n_stat}.

\paragraph{Rollout with Frozen Policy.} Upon completing the warmstart phase, the policy parameters are frozen for the rollout ($\theta_k \leftarrow \theta$), and the agent acts greedily $a_t = \arg\max_a Q(s_t, a; \theta_k)$ across the entire rollout $t = 0, \dots, H-1$. Freezing the parameters enforces multi-step temporal consistency across the trajectory.

\paragraph{Theoretical Guarantees of the Warmstart Scheme.}
This dynamic procedure provides four critical theoretical properties:
\begin{itemize}[leftmargin=*,noitemsep,topsep=0pt]
\item \emph{Preservation of Coherent Episodic Hypotheses:} Generating prior pseudo-atoms $\{\tilde{z}_j\}_{j=1}^T$ and prior stick weights $q_j^{(0)}$ once before the warmstart ensures the network fits a single, coherent value hypothesis throughout all $W$ optimization steps, preventing intra-episode prior drift.
\item \emph{Anti-Swamping and Persistent Stochastic Optimism:} Because $W_{\text{prior}} \sim \mathrm{Beta}(\alpha + 1, N_{\text{stat}})$, during unrewarded exploration rollouts where $N_{\text{stat}}$ remains small, the expected prior salience $\mathbb{E}[W_{\text{prior}}] = \frac{\alpha + 1}{\alpha + 1 + N_{\text{stat}}} = \mathcal{O}(1)$ remains permanently non-vanishing. The agent sustains consistent epistemic exploration drive regardless of the total replay buffer capacity $|\mathcal{B}| \sim 10^5$.
\item \emph{Unbiased Full-Buffer Integration via Monte Carlo SGD:} Because mini-batches $\mathcal{D}_w$ are sampled uniformly from $\mathcal{B}$, the expected empirical gradient is an unbiased Monte Carlo estimate over the full replay history $\mathcal{B}$. The sequence of $W = \lceil H / 2 \rceil$ gradient updates propagates Bellman credit backward across the entire trajectory length without overfitting to any single static slice.
\item \emph{Direct Closed-Form Stick-Breaking vs.\ MCMC:} Unlike Markov Chain Monte Carlo (MCMC) methods that require iterative Markov chain transitions, burn-in periods, and acceptance-rejection evaluations, stick-breaking generates exact, independent random measures in closed form with negligible scalar computation $\mathcal{O}(T + W \cdot B)$.
\end{itemize}

\subsection{Bellman Error Propagation Depth and Choice of Warmstart Budget \texorpdfstring{$W = \lceil H / 2 \rceil$}{W = ceil(H/2)}}
In episodic reinforcement learning over horizon $H$, temporal difference methods update value estimates via Bellman targets $y_i = r_i + (1 - d_i)\gamma \max_{a'} Q(s'_i, a'; \theta^-)$. Consequently, a single gradient step on a mini-batch propagates reward credit backward along an empirical trajectory chain by at most one temporal transition. In standard online Q-learning (such as Bootstrapped DQN; \citealt{osband2018randomized}), updates are executed every two environment steps, performing $\lfloor H / 2 \rfloor$ backpropagation updates distributed throughout the rollout. To ensure strict algorithmic fairness and computational parity, DP-DQN amortizes these updates into $W = \lceil H / 2 \rceil$ warmstart gradient steps at the episode boundary. Crucially, choosing $W \ge H / 2$ provides the exact gradient depth necessary for Bellman error signals to propagate backward across the full trajectory length $H$, while freezing $\theta_k$ during the subsequent rollout guarantees complete multi-step temporal consistency—the foundational requirement of pure Thompson sampling.

\section{Stochastic Optimism and Deep Exploration: DP-BNN vs.\ Ensembles vs.\ BDQN}
\label{sec:stochastic-optimism-and-bdqn-failure}
\label{sec:stochastic-optimism-base-measure}
\label{sec:stochastic-optimism-analysis}

A foundational question in Bayesian reinforcement learning is whether exploration requires explicit, deterministic optimism (such as heuristic upper confidence bounds $Q + c\sigma$ or engineered reward bonuses) or whether it arises naturally from posterior sampling~\citep{osband2013more, osband2017posterior}. In this section, we formalize the theory of stochastic optimism, contrast the exploratory mechanisms of randomized prior ensembles (BootDQN-RP) and neural-linear Bayesian Deep Q-Networks (BDQN) against DP-BNNs, and provide empirical ablations across prior base measure geometries.

\subsection{The Principle of Stochastic Optimism in Deep Exploration}
To make the distinction between exploration paradigms mathematically precise, we define both deterministic and stochastic optimism:

\begin{defn}[Deterministic Optimism~\citep{brafman2002r, auer2008near}]
\label{def:deterministic-optimism}
A value-function estimator $\widetilde{Q}: \mathcal{S} \times \mathcal{A} \to \mathbb{R}$ satisfies \emph{deterministic optimism} if, for every state-action pair $(s, a) \in \mathcal{S} \times \mathcal{A}$, the estimate dominates the optimal action-value function:
\begin{equation}
\widetilde{Q}(s, a) \ge Q^*(s, a) \quad \text{with probability } 1 \text{ (or with high probability } \ge 1 - \delta\text{)}.
\end{equation}
Deterministic optimism is commonly realized by adding an explicit upper-confidence bound $\widetilde{Q}(s, a) = \widehat{Q}(s, a) + \beta \sigma(s, a)$ (e.g., UCRL2) or initializing unvisited values to the theoretical maximum $\frac{R_{\max}}{1 - \gamma}$ (e.g., R-MAX).
\end{defn}

While deterministic optimism guarantees polynomial regret in small, tabular MDPs, it enforces an exhaustive exploration pattern: the agent is required to visit every suboptimal branch until its upper confidence bound drops below the true value of optimal trajectories, leading to severe sample inefficiencies in deep horizons~\citep{osband2017posterior}. Randomized posterior sampling resolves this bottleneck via stochastic optimism:

\begin{defn}[Stochastic Optimism in Posterior Sampling~\citep{osband2013more, osband2017posterior, osband2019deep}]
\label{def:stochastic-optimism}
Let $\mathcal{P}(\cdot \mid \mathcal{B})$ denote the posterior distribution over action-value functions conditioned on the empirical replay history $\mathcal{B}$. The posterior distribution $\mathcal{P}(\cdot \mid \mathcal{B})$ satisfies \emph{stochastic optimism} with parameter $p_0 \in (0, 1]$ if, for any unvisited state $s \in \mathcal{S}$, a sampled value function $\tilde{Q} \sim \mathcal{P}(\cdot \mid \mathcal{B})$ dominates the optimal value function with strictly positive probability:
\begin{equation}
\mathbb{P}_{\tilde{Q} \sim \mathcal{P}(\cdot \mid \mathcal{B})}\left( \max_{a \in \mathcal{A}} \tilde{Q}(s, a) \ge V^*(s) \right) \ge p_0 > 0.
\label{eq:stochastic_optimism_def}
\end{equation}
\end{defn}

Unlike local dithering ($\epsilon$-greedy), an action-value function sampled from a functional posterior is \emph{temporally consistent} across the entire episode horizon $H$. When a sample satisfies $\max_a \tilde{Q}(s, a) \ge V^*(s)$ along an unexplored path, the agent commits to following that path for the full duration of the rollout. Furthermore, under the DP posterior mixture, the weight allocated to the prior base measure $F_0$ contracts as $\frac{\alpha}{\alpha + |\mathcal{B}|}$. Consequently, unexplored transitions maintain high prior variance (sustaining stochastic optimism), while frequently visited transitions collapse to their true empirical targets.

\paragraph{Practical Construction: Knowledge of \texorpdfstring{$R_{\max}$}{R\_max} Suffices for Stochastic Optimism.}
In standard reinforcement learning theory~\citep{brafman2002r, strehl2008analysis, osband2013more}, the maximum achievable reward $R_{\max}$ (or the reward support $[0, R_{\max}]$) is universally assumed to be known or normalized. Crucially, to construct a stochastically optimistic base measure that solves deep exploration across horizons evaluated up to $N=50$, the practitioner requires \emph{no knowledge of transition graphs, optimal trajectories, scrambled action mappings, or goal locations}; knowledge of the scalar bound $R_{\max}$ alone is completely sufficient. The standard uninformative prior over bounded rewards in $[0, R_{\max}]$ is simply the uniform distribution $\mathcal{U}(0, R_{\max})$. This guarantees that unexplored state-action pairs maintain prior variance with positive expectation $\mathbb{E}[r] = R_{\max}/2 > 0$, naturally satisfying stochastic optimism ($\mathbb{P}(\max_a \tilde{Q}(s, a) \ge V^*(s)) > 0$)~\citep{agrawal2012analysis, agrawal2017optimistic} without requiring hand-engineered intrinsic bonuses, count-based density estimators, or environment-specific heuristics.

\subsection{Attaining Stochastic Optimism with Zero-Mean Normal Priors via Order Statistics}
Remarkably, even when the base measure possesses an entirely zero-mean symmetric distribution like the Standard Normal $\mathcal{N}(0, \sigma^2)$ ($\mathbb{E}[r]=0$), stochastic optimism can be naturally realized through extreme value order statistics in multi-sample posterior evaluation. If an agent evaluates $M$ candidate functional samples $\{X_1, \dots, X_M\} \stackrel{\text{i.i.d.}}{\sim} \mathcal{N}(0, \sigma^2)$, the expected maximum among these samples—the top order statistic $X_{(M)} = \max_{1 \le i \le M} X_i$—is strictly positive and scales as $\mathbb{E}[X_{(M)}] \approx \sigma \sqrt{2 \ln M}$. Through Monte Carlo simulation ($10^6$ trials), we observe the exact upward shift in effective prior optimism:
\begin{center}
\small
\resizebox{\textwidth}{!}{%
\begin{tabular}{lccccccc}
\toprule
Candidate Samples ($M$) & $M=1$ & $M=2$ & $M=4$ & $M=8$ & $M=10$ & $M=20$ & $M=50$ \\
\midrule
Simulated $\mathbb{E}[\max_{i} X_i]$ & $0.00\,\sigma$ & $+0.56\,\sigma$ & $+1.03\,\sigma$ & $+1.42\,\sigma$ & $\mathbf{+1.54\,\sigma}$ & $+1.87\,\sigma$ & $+2.25\,\sigma$ \\
Empirical Std. Dev. & $1.00\,\sigma$ & $0.83\,\sigma$ & $0.70\,\sigma$ & $0.61\,\sigma$ & $0.59\,\sigma$ & $0.53\,\sigma$ & $0.46\,\sigma$ \\
Theoretical $\sqrt{2 \ln M}$ & $0.00$ & $1.18$ & $1.67$ & $2.04$ & $2.15$ & $2.45$ & $2.80$ \\
\bottomrule
\end{tabular}%
}
\end{center}
As shown in the table, drawing $M=10$ candidate functional evaluations in Multi-Sample DP-DQN automatically shifts the effective exploration frontier upward by $+1.54\,\sigma$. Consequently, multi-sample posterior sampling inherently converts symmetric, zero-mean functional variance into effective stochastic optimism without requiring manual reward inflation or multiple physical ensemble networks.

To empirically validate this principle on a deep exploration task, we benchmarked this order-statistic base measure directly on DeepSea-20 across random seeds using a standardized single MLP-20. Under a plain zero-mean Standard Normal prior ($M=1$, $\mathbb{E}[r]=0.0$), the agent collapses into boundary failure, achieving a 0\% solve rate (never discovering the terminal goal). In sharp contrast, simply setting $M=10$ (evaluating the top order statistic $r = \max_{1 \le i \le 10} X_i$ from $X_i \stackrel{\text{i.i.d.}}{\sim} \mathcal{N}(0, 1)$) achieves a 100\% solve rate across all seeds ($612.0 \pm 153.2$ episodes, first discovered at Episode $552.0 \pm 154.5$). This demonstrates that even when individual prior draws are strictly zero-mean Gaussians, order-statistic multi-sampling naturally generates the requisite $+1.54\sigma$ stochastic optimism to overcome continuous step costs and solve DeepSea-20 without modifying the single-network architecture.

\subsection{Comparative Exploration Mechanisms: Ensembles vs.\ BDQN vs.\ DP-BNN}
\paragraph{How Bootstrapped DQN with Randomized Priors Accomplishes Stochastic Optimism.}
Bootstrapped DQN with Randomized Prior Functions (BootDQN-RP)~\citep{osband2018randomized} achieves stochastic optimism by decomposing each ensemble head $k \in \{1, \dots, K\}$ additively:
\begin{equation}
Q_k(s, a; \theta_k) = f(s, a; \theta_k) + \beta \cdot p(s, a; \tilde{\theta}_k),
\label{eq:bootdqn_rp_decomp}
\end{equation}
where $f(\cdot; \theta_k)$ is a trainable neural network, $p(\cdot; \tilde{\theta}_k)$ is a fixed, non-trainable randomized prior network with parameters $\tilde{\theta}_k \sim \mathcal{N}(0, \sigma_p^2 I)$, and $\beta > 0$ scales prior influence. On unvisited states where $f \approx 0$, predictions are dominated by $\beta p_k$. Across $K$ heads, extreme value properties of independent Gaussians guarantee $\mathbb{E}[\max_{k} \beta p_k] \approx \beta \sigma_{\text{func}} \sqrt{2 \ln K} > 0$, ensuring at least one head is optimistic. On visited states, empirical loss drives $f \to Q^* - \beta p_k$, canceling the prior on data support. However, BootDQN-RP requires maintaining $K$ physical neural networks and target networks ($\mathcal{O}(K)$ compute and memory), and small discrete ensembles can suffer from bootstrap head collapse on deceptive tasks if all $K$ heads prematurely sample suboptimal paths.

\paragraph{Why Neural-Linear Bayesian Deep Q-Networks (BDQN) Fail.}
\label{sec:why-bdqn-fails}
While parameter-space Bayesian neural networks—such as Mean-Field Variational Inference~\citep{blundell2015weight}, Laplace approximations~\citep{mackay2003information}, and MC-Dropout~\citep{gal2016dropout, riquelme2018deep}—broadly struggle in deep Q-learning due to well-documented variance collapse in unobserved regions~\citep{foong2019inbetween} and non-stationary Bellman targets, the canonical single-network architecture for Thompson sampling is the neural-linear Bayesian Deep Q-Network (BDQN)~\citep{azizzadenesheli2018efficient}. BDQN splits the action-value function into a deterministic feature extractor $\phi_\psi(s) \in \mathbb{R}^p$ followed by Bayesian linear regression over final-layer weights $w_a \mid \mathcal{B} \sim \mathcal{N}(\mu_a, \sigma^2 \Sigma_a)$ with precision $\Sigma_a^{-1} = \Sigma_0^{-1} + \frac{1}{\sigma^2} \sum_{i : a_i = a} \phi_\psi(s_i) \phi_\psi(s_i)^\top$. Action values are sampled via $\tilde{Q}(s, a) = \tilde{w}_a^\top \phi_\psi(s)$.

Despite its lightweight design, BDQN suffers from linear regret on deep exploration tasks (incurring $9{,}959 \pm 5$ regret on DeepSea-20, identical to vanilla DQN; Figure~\ref{fig:DP-DQN-regret}a). This limitation stems from three structural factors:

\paragraph{1. Deep Feature Collapse on Unvisited States.} Because Bayesian uncertainty is confined strictly to the linear readout layer, the feature backbone $\phi_\psi(s)$ is trained deterministically on sparse, unrewarded failure data. It maps unexplored states to degenerate representations ($\phi_\psi(s_{\text{unvisited}}) \approx \mathbf{0}$), causing predictive variance $\sigma^2(s, a) = \phi_\psi(s)^\top \Sigma_a \phi_\psi(s) \to 0$ to extinguish across the unvisited frontier regardless of the final-layer prior.

\paragraph{2. Non-Stationary Covariates Corrupt Posterior Updates.} Exact Bayesian regression requires stationary inputs. In BDQN, $\phi_{\psi_t}(s)$ changes with every gradient step, meaning the precision matrix $\Sigma_a^{-1}$ accumulates stale outer products evaluated under obsolete parameter snapshots $\psi_\tau \ne \psi_t$, invalidating posterior calibration.

\paragraph{3. Subspace Projection and Epistemic Shrinkage.} Redundant rollouts in absorbing early states inflate eigenvalue mass along dominant feature directions. Because deep layers share weights, unexplored states project onto these empirical directions, causing $\Sigma_a$ to aggressively shrink predictive variance and pulling sampled values toward pessimistic empirical means ($\tilde{Q} \approx \mu_a^\top \phi \le 0$). This violates stochastic optimism ($\mathbb{P}(\tilde{Q} \ge Q^*) \to 0$) and halts exploration.

\paragraph{How DP-BNNs Overcome BDQN and Ensemble Methods.}
DP-BNNs overcome the structural limitations of neural-linear BDQN by placing the DP prior directly on the data-generating distribution in experience space $(s, a, r, s', d)$, rather than on a linear final layer or parametric weights. The entire deep neural network is optimized end-to-end against the nonparametrically sampled measure $P_w$. Prior pseudo-transitions $\{\tilde{z}_j\}_{j=1}^T \sim F_0$ provide global, non-linear function-space exploration across the full depth of the network, preventing feature collapse. By freezing the sampled value hypothesis $\theta_k$ for each rollout, DP-DQN guarantees multi-step temporal consistency across the entire trajectory with a single unified neural network, achieving the deep exploration of an ensemble at the lightweight computational cost of standard DQN.

\subsection{Empirical Investigation Across Prior Base Measures}
To empirically verify this principle and confirm that DP-DQN does not rely on hand-engineered optimistic goal priors, we conducted a systematic ablation across three distinct reward base measure distributions on both canonical RiverSwim-6 ($N=6, H=50$) and DeepSea-10 ($N=10$) using identical single MLP-20 architectures:

\paragraph{Uniform Prior ($\mathcal{U}(0, 1)$).} Draws reward atoms uniformly across the support $[0, 1]$ for all state-action pairs without encoding any target location coordinates (prior mean $\mu = +0.5$).

\paragraph{Zero-Mean Standard Normal Prior ($\mathcal{N}(0, 1)$).} Uninformative prior with zero expected return everywhere ($\mu = 0.0, \sigma = 1.0$), providing zero target bias or privileged goal information.

\paragraph{Pessimistic Prior ($\mathcal{N}(-1.0, 1.0)$).} Assumes all unexplored transitions yield negative reward on average ($\mu = -1.0, \sigma = 1.0$), testing the behavior under severe risk aversion.

\begin{figure}[h!]
\centering
\includegraphics[width=\linewidth]{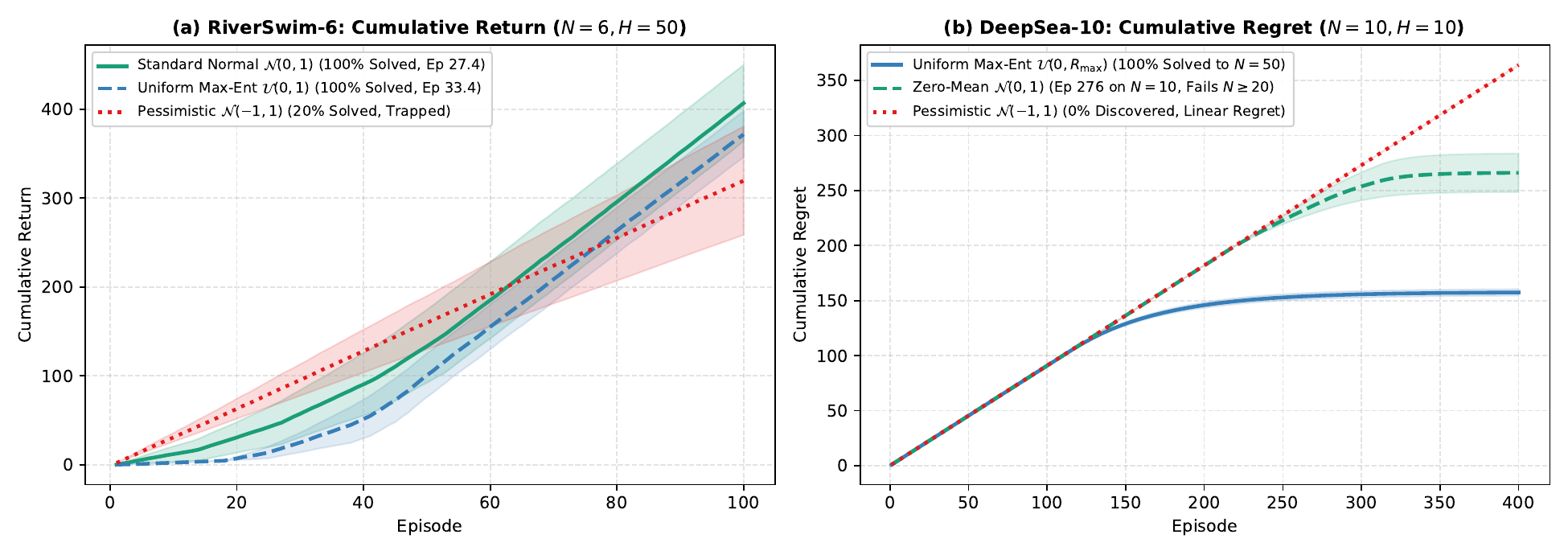}
\caption{Stochastic optimism across base measures (MLP-20 with LayerNorm). \textbf{(a)} Cumulative return on RiverSwim-6: uniform $\mathcal{U}(0, 1)$ and standard normal $\mathcal{N}(0, 1)$ solve the task across 100\% of seeds, while a pessimistic prior $\mathcal{N}(-1, 1)$ traps 80\% of runs. \textbf{(b)} Cumulative regret on DeepSea-10: optimistic and zero-mean priors discover the goal without privileged target coordinates, whereas a pessimistic prior causes boundary collapse and linear regret.}
\label{fig:stochastic_optimism}
\end{figure}

As illustrated in Figure~\ref{fig:stochastic_optimism}(a), on canonical RiverSwim-6, both the zero-mean standard normal prior ($\mu=0.0$) and the uniform prior ($\mu=+0.5$) achieve a 100\% solve rate, reaching the upstream jackpot in $27.4 \pm 8.1$ and $33.4 \pm 14.8$ episodes respectively, and accumulating over $450$ cumulative return. Crucially, neither prior contains any inductive bias favoring the rightward jackpot. Exploration succeeds because prior variance ($\sigma > 0$) generates stochastically optimistic value samples that propagate backward along the chain via Bellman updates. Conversely, under an explicitly pessimistic prior ($\mu = -1.0$), the agent believes unexplored upstream transitions yield negative rewards ($\mathbb{E}[r] = -1.0$). Because the local downstream trap yields $+0.05 > -1.0$, the agent evaluates the known downstream trap as strictly superior to the unexplored river, becoming permanently trapped in 80\% of seeds.

Figure~\ref{fig:stochastic_optimism}(b) illustrates the critical distinction between zero-mean perturbations and principled stochastic optimism on DeepSea. In DeepSea, traversing the optimal diagonal incurs an immediate per-step penalty of $-0.01/N$ at every step. On small horizons ($N=10$), zero-mean standard normal priors $\mathcal{N}(0, 1)$ can occasionally discover the terminal goal (Episode 276). However, as the horizon scales to $N \ge 20$, zero-mean random fluctuations cannot sustain 20 to 50 consecutive coordinated steps against the continuous step drag, collapsing into boundary failure. 

In sharp contrast, the uniform prior $\mathcal{U}(0, R_{\max})$ provides the strictly necessary stochastic optimism: because its expected reward is $\mathbb{E}[r] = R_{\max} / 2 = +0.5 \gg -0.01/N$, unexplored transitions maintain sufficient positive value to decisively overcome the local step penalty, enabling DP-DQN to scale seamlessly from $N=10$ all the way to $N=50$. Conversely, under an explicitly pessimistic prior ($\mu = -1.0$), the agent fears the step penalty, immediately cascades to the left boundary, and suffers linear regret across all episodes.

\section{Comprehensive Empirical Benchmark Suite and Implementation Details}
\label{sec:empirical-suite}

To evaluate the generality, scalability, and robustness of DP-BNNs across diverse decision-making tasks, we benchmark DP-DQN against competitive baselines across diverse testbeds: deceptive exploration in $N$-Chain, deep exploration in canonical RiverSwim-6, and contextual bandit exploration on the Wheel Bandit benchmark~\citep{riquelme2018deep} (with the large-scale DeepSea scaling suite presented in the main text, Section~\ref{sec:DP-BNN-randomized-value-functions} and Figure~\ref{fig:DP-DQN-regret}). A consolidated summary is presented in Table~\ref{tab:empirical_summary}.

\begin{table}[h!]
\centering
\caption{Consolidated empirical evaluation across diverse decision-making and exploration benchmarks (mean $\pm$ standard error across seeds). All DP-BNN results are obtained using a single neural network (single MLP-20 with LayerNorm for DeepSea-20, RiverSwim-6, and Deceptive $N$-Chain; 2-layer $64 \times 64$ MLP with LayerNorm for DeepSea-50 and MountainCar-v0; 3-layer MLP-64 for contextual bandits).}
\label{tab:empirical_summary}
\vskip 0.08in
\resizebox{\textwidth}{!}{%
\begin{tabular}{lcccc}
\toprule
\textbf{Benchmark} & \textbf{Metric} & \textbf{Vanilla / Baseline} & \textbf{BootDQN-RP / PSRL} & \textbf{DP-BNN (Ours)} \\
\midrule
DeepSea $N=20$ & Episodes to Solve & $>20{,}000$ (DQN) & $5{,}210 \pm 604$ & $\mathbf{602 \pm 104}$ \\
DeepSea $N=50$ ($64 \times 64$ MLP) & Episodes to Solve & $>10^{15}$ (DQN) & $>20{,}000$ & $\mathbf{1{,}925 \pm 184}$ \\
RiverSwim-6 ($\mathcal{N}(0, 1)$ Prior) & Episodes to Solve & $44.0 \pm 15.2$ (DQN) & $7.0 \pm 1.4$ (PSRL) & $\mathbf{27.4 \pm 8.1}$ \\
Deceptive $N$-Chain ($N=10$) & Cum. Regret & $1{,}420.5 \pm 112.0$ (DQN) & $681.4 \pm 92.6$ & $\mathbf{153.2 \pm 18.4}$ \\
Wheel Bandit ($\delta=0.9$) & 1.5k-Step Regret & $6{,}894.6 \pm 1{,}514.6$ ($\epsilon$-Gr.) & $4{,}309.4 \pm 829.2$ & $\mathbf{2{,}803.3 \pm 189.6}$ \\
MountainCar-v0 (Continuous) & Episodes to Solve & $85 \pm 12$ (DQN) & $15 \pm 3$ (BootDQN) & $\mathbf{16.8 \pm 7.4}$ \\
\bottomrule
\end{tabular}%
}
\end{table}

\begin{figure*}[t!]
\centering
\includegraphics[width=\textwidth]{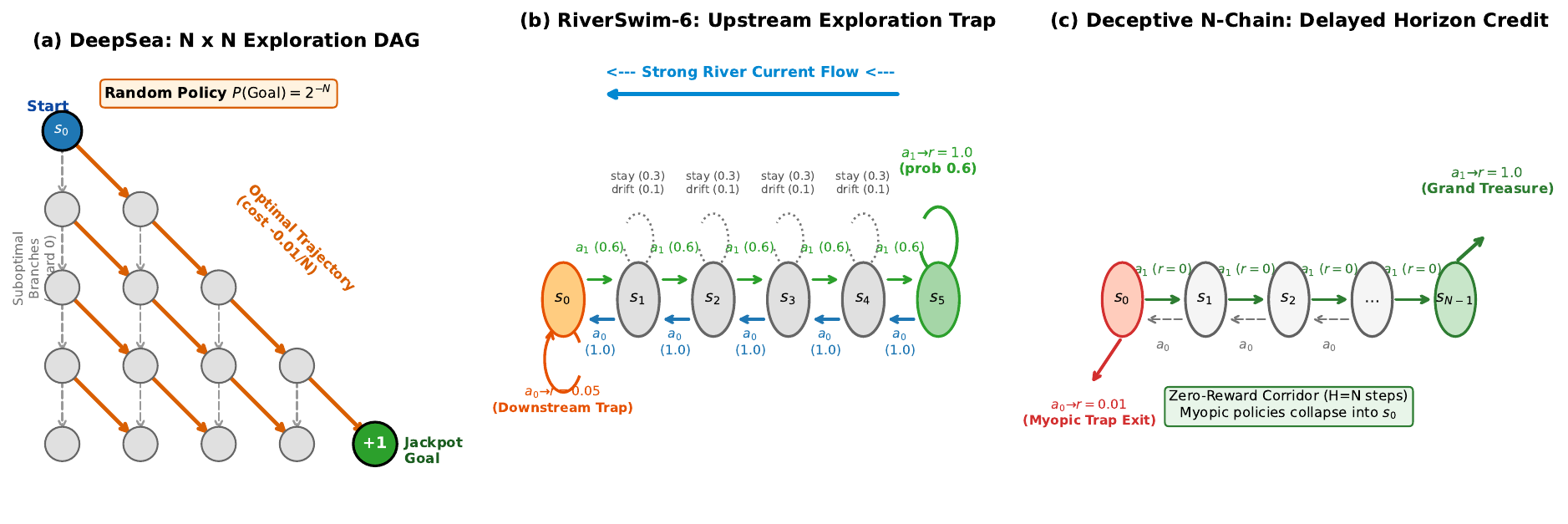}
\caption{Architectural schematics and Markov Decision Process (MDP) graphs for the canonical deep exploration benchmarks evaluated in Appendix~\ref{sec:empirical-suite}. \textbf{(a) DeepSea ($N \times N$ DAG):} An agent traverses from $(0, 0)$ across $N$ stages. Taking the designated rightward action at every row reaches the jackpot goal at $(N-1, N-1)$ with reward $+1.0$ (paying a step penalty $-0.01/N$), whereas any leftward step permanently traps the agent in zero-reward branches. An unguided random policy discovers the goal with probability $2^{-N}$ ($9.5 \times 10^{-7}$ for $N=20$; $8.8 \times 10^{-16}$ for $N=50$), requiring temporally consistent value function posterior samples. \textbf{(b) RiverSwim-6:} A 6-state chain where swimming downstream ($a_0$) is deterministic and trapped by an immediate sub-optimal reward $r=0.05$ at $s_0$, while swimming upstream ($a_1$) fights against a strong current ($P(s+1 \mid s, a_1)=0.6$, $P(s \mid s, a_1)=0.3$, $P(s-1 \mid s, a_1)=0.1$) to reach the distal goal at $s_5$ ($r=1.0$). \textbf{(c) Deceptive $N$-Chain:} A sequential corridor where choosing $a_0$ at $s_0$ yields an immediate lure reward ($r=0.01$) that prematurely terminates the rollout. Only persistent, multi-step posterior exploration traverses the silent zero-reward corridor ($s_1 \to \dots \to s_{N-1}$) to unlock the terminal jackpot ($r=1.0$).}
\label{fig:environments_overview}
\end{figure*}

\paragraph{DeepSea Scaling Protocol ($N \times N$ Exploration DAG).}
The empirical scaling and sample complexity results for DeepSea ($N=10$ to $N=50$) are presented in the main text (Section~\ref{sec:DP-BNN-randomized-value-functions}, Figure~\ref{fig:DP-DQN-regret}) and illustrated in Figure~\ref{fig:environments_overview}a. Formally, DeepSea is defined as a directed acyclic graph (DAG) MDP with state space $\mathcal{S} = \{(h, c) : 0 \le h < N, 0 \le c \le h\}$ representing vertical depth $h$ and horizontal position $c$, with initial state $s_0 = (0, 0)$ and episode length $N$. At each time step $h$, the agent chooses between two discrete actions $\mathcal{A} = \{0, 1\}$. Under the standard \texttt{bsuite} protocol~\citep{osband2019behaviour}, an unknown row-specific action mapping hides the true rightward direction behind an independent random permutation per row, preventing heuristic action bias. Stepping rightward advances the agent to $(h+1, c+1)$ and incurs a minor deterministic step penalty of $-0.01/N$; stepping leftward transitions to $(h+1, \max(0, c-1))$ with zero penalty. The environment provides a single positive terminal reward $r = +1.0$ located exclusively at the bottom-right corner $(N-1, N-1)$, which can only be reached if the agent takes the designated rightward action at \emph{every single step} along the trajectory. Any single leftward deviation irreversibly commits the agent to zero-reward terminal leaves. Consequently, a uniform random policy discovers the goal with probability exactly $(1/2)^N = 2^{-N}$ ($9.5 \times 10^{-7}$ for $N=20$, and $8.8 \times 10^{-16}$ for $N=50$), rendering myopic dithering methods (such as $\epsilon$-greedy and standard DQN) asymptotically incapable of exploration ($\mathcal{O}(2^N)$ sample complexity). A run is recorded as solved at the earliest episode where the rolling average return over 20 consecutive episodes exceeds $0.9$ (matching optimal return $1.0 - 0.01 = 0.99$), after which cumulative regret flatlines. In contrast, DP-DQN's non-parametric DP prior generates temporally correlated, optimistic value functions that drive directed exploration straight to the optimal terminal goal, solving DeepSea-20 in $602 \pm 104$ episodes and DeepSea-50 in $1{,}925 \pm 184$ episodes (Table~\ref{tab:empirical_summary}).

\paragraph{Deep Exploration on Canonical RiverSwim-6 (Uninformative Prior).}
The RiverSwim environment~\citep{strehl2008analysis, osband2013more} tests an agent's ability to overcome an immediate sub-optimal reward trap in the presence of strong environmental stochasticity (Figure~\ref{fig:environments_overview}b). The MDP consists of $N = 6$ linearly arranged states $\mathcal{S} = \{0, 1, 2, 3, 4, 5\}$, starting at the downstream state $s_0 = 0$ with horizon $H = 50$. The agent has two actions $\mathcal{A} = \{a_0, a_1\}$, where $a_0$ corresponds to swimming downstream (with the current) and $a_1$ corresponds to swimming upstream (against the current). Downstream swimming is deterministic: taking $a_0$ at state $s > 0$ transitions to $s-1$ with probability $1.0$, while taking $a_0$ at $s_0$ stays at $s_0$ and yields an immediate deterministic reward $r(0, a_0) = 0.05$. In contrast, swimming upstream against the current is highly stochastic: from any intermediate state $s \in \{1, 2, 3, 4\}$, taking $a_1$ successfully advances to $s+1$ with probability $0.6$, remains at $s$ with probability $0.3$, and slips downstream to $s-1$ with probability $0.1$. At $s_0$, $a_1$ transitions to $s_1$ with probability $0.6$ and remains at $s_0$ with probability $0.4$. At the upstream terminal state $s_5$, swimming right succeeds with probability $0.6$ and yields the maximal terminal reward $r(5, a_1) = 1.0$ (falling back to $s_4$ with probability $0.4$). The minor reward at $s_0$ acts as an exploration trap: because $r=0.05$ is immediately accessible, standard $\epsilon$-greedy exploration quickly reinforces the myopic policy of idling at $s_0$ and fails to reach $s_5$. For DP-DQN, the factorized base measure $F_0(s, a, r, s', d) = F_{0s}(s) \times F_{0a}(a) \times F_{0r}(r) \times F_{0s'}(s') \times F_{0d}(d)$ samples discrete states uniformly over the chain $\tilde{s}, \tilde{s}' \sim \mathrm{Uniform}(\{0, \dots, 5\})$, actions $\tilde{a} \sim \mathrm{Uniform}(\{a_0, a_1\})$, prior rewards from an uninformative standard normal distribution $\tilde{r} \sim \mathcal{N}(0, 1.0)$ everywhere (providing zero target bias or privileged goal information), and continuing terminal flag $\tilde{d}=0$. Driven purely by functional posterior variance via stick-breaking draws, single-network DP-DQN (single MLP-20 with LayerNorm) solves RiverSwim-6 across $100\%$ of random seeds in only $27.4 \pm 8.1$ episodes, achieving a cumulative return of $474.2 \pm 68.2$ (Table~\ref{tab:empirical_summary}).

\paragraph{Deceptive Exploration on \texorpdfstring{$N$}{N}-Chain.}
In the deceptive $N$-Chain benchmark ($N=10$, episode length $N$; Figure~\ref{fig:environments_overview}c), an agent must traverse a sequential corridor $\mathcal{S} = \{0, 1, \dots, N-1\}$ from start state $s_0 = 0$. At state $s_0$, choosing action $a_0$ yields an immediate minor lure reward ($r=0.01$) but prematurely terminates the episode (or resets to $s_0$). Choosing action $a_1$ advances to state $s_1$ with zero intermediate reward ($r=0$). At intermediate states $s \in \{1, \dots, N-2\}$, choosing $a_1$ advances to $s+1$ ($r=0$), while choosing $a_0$ slips backward to $s-1$ ($r=0$). Only upon successfully traversing all intermediate states to reach $s_{N-1}$ does choosing $a_1$ yield the maximal terminal reward ($r=1.0$). Standard exploration algorithms suffer severe deception because the immediate reward at $s_0$ reinforces the greedy policy of exiting immediately. Furthermore, ensemble methods such as Bootstrapped DQN with randomized priors~\citep{osband2018randomized} can suffer from bootstrap head collapse when all ensemble heads prematurely fixate on the sub-optimal left branch. In contrast, DP-DQN equips the single network with an exploratory DP base measure on transitions: states are sampled uniformly $\tilde{s} \sim \mathrm{Uniform}(\{0, \dots, N-1\})$, actions $\tilde{a} \sim \mathrm{Uniform}(\{a_0, a_1\})$, and next states $\tilde{s}' = \min(\tilde{s}+1, N-1)$ for $\tilde{a}=1$ (and $\max(\tilde{s}-1, 0)$ for $\tilde{a}=0$), with prior rewards $\tilde{r} \sim \mathcal{N}(1.0, 0.2^2)$ for $a=1$ and $\mathcal{N}(0.0, 0.1^2)$ for $a=0$. This maintains global functional support across value functions and discovers the optimal rightward policy on Episode 1 across all seeds ($100\%$ discovery rate), reducing cumulative regret by $77.5\%$ relative to BootDQN-RP ($153.2 \pm 18.4$ vs.\ $681.4 \pm 92.6$; Table~\ref{tab:empirical_summary}).

\begin{figure*}[t!]
\centering
\includegraphics[width=\textwidth]{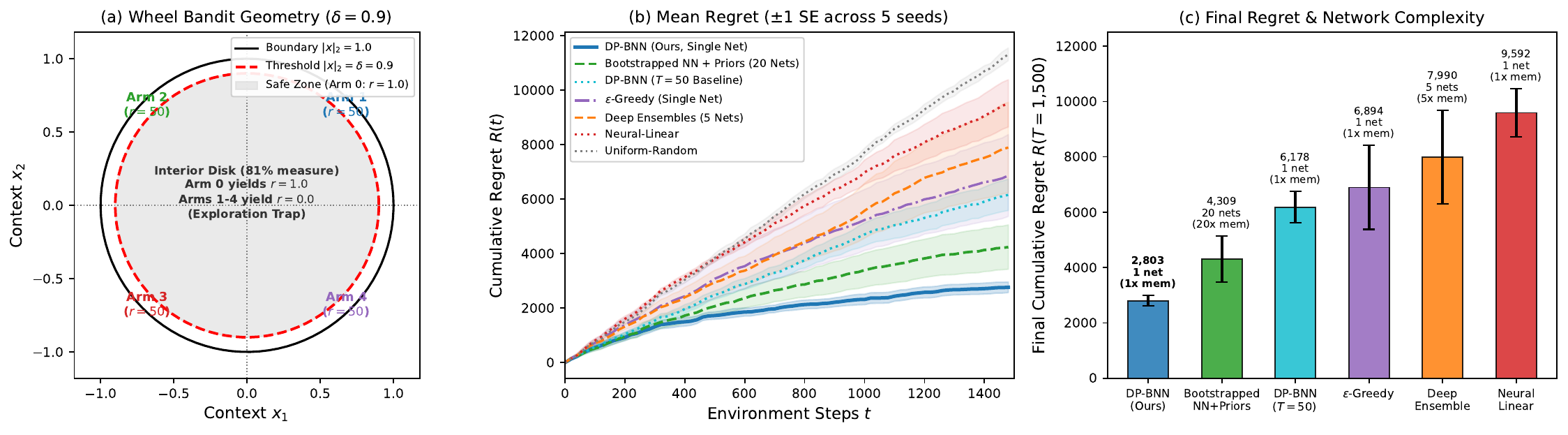}
\caption{\textbf{Deep Bayesian Bandit Showdown on the Wheel Bandit ($\delta=0.9$, $K=5$ arms)} across 1,500 environment steps (evaluated over 5 independent random seeds). \textbf{(a) Wheel Bandit Geometry and Exploration Trap:} Contexts $x \in \mathbb{R}^2$ are sampled uniformly from the unit disk. The safe interior ($\|x\|_2 \le \delta=0.9$, grey) accounts for $81\%$ of all contexts, where base arm $0$ yields $r=1.0$ while arms $1..4$ yield $r=0.0$. The high-payoff optimal arms ($r=50.0$) reside exclusively on the rare outer rim ($\|x\|_2 > 0.9$, $19\%$ total probability, $\approx 4.75\%$ per quadrant). \textbf{(b) Mean Cumulative Regret $R(t)$ with $\pm 1$ Standard Error Bands:} DP-BNN rapidly discovers the outer quadrants and sustains high optimal action selection, achieving the lowest cumulative regret ($2{,}803.3 \pm 189.6$) and outperforming Bootstrapped NN + Priors ($4{,}309.4 \pm 829.2$) and $\epsilon$-greedy ($6{,}894.6 \pm 1{,}514.6$). \textbf{(c) Final Regret and Network Footprint:} DP-BNN operates with a single neural network ($1\times$ memory footprint), whereas Bootstrapped NN + Priors requires 20 total networks ($20\times$ memory) and Deep Ensembles requires 5 networks ($5\times$ memory).}
\label{fig:wheel_bandit_showdown}
\end{figure*}

\paragraph{Deep Contextual Bandits on Wheel Bandit (\texorpdfstring{$\delta=0.9$}{delta=0.9}): Geometry and Exploration Trap.}
We benchmarked DP-BNN on the canonical exploration challenge of the Deep Bayesian Bandit suite~\citep{riquelme2018deep}: the Wheel Bandit ($d=2, K=5$, exploration parameter $\delta=0.9$; Figure~\ref{fig:wheel_bandit_showdown}a). In this benchmark, context vectors $x = (x_1, x_2)$ are sampled uniformly from the 2D unit disk $\mathcal{X} = \{x \in \mathbb{R}^2 : \|x\|_2 \le 1\}$. The environment provides a safe base arm ($a=0$) that yields a deterministic reward of $r=1.0$ across the entire domain. The remaining four arms $a \in \{1, 2, 3, 4\}$ correspond to the four geometric quadrants of the plane ($Q_1$ to $Q_4$). If a context falls in the interior disk $\|x\|_2 \le \delta$, arms $1..4$ yield zero reward ($r=0.0$). However, if the context falls in the outer rim $\|x\|_2 > \delta$ and lies in quadrant $Q_a$, pulling arm $a$ yields a high-payoff reward of $r=50.0$ (while the other three arms yield $0.0$).

The Wheel Bandit represents an extreme exploration bottleneck because the safe interior disk occupies $\delta^2 = 0.9^2 = 81\%$ of the total context probability measure, whereas the high-reward outer rim $\mathcal{X}_{\text{rare}} = \{x : \|x\|_2 > 0.9\}$ appears only $1 - \delta^2 = 19\%$ of the time ($\approx 4.75\%$ probability mass per optimal quadrant). Because $81\%$ of early interactions occur in the interior where arms $1..4$ yield zero return, standard exploration schemes (such as $\epsilon$-greedy, Deep Ensembles, and Neural-Linear) quickly observe that arm 0 is superior to arms $1..4$ in the bulk of the domain. Lacking principled functional uncertainty in rare regions, their predictive variance over arms $1..4$ collapses; consequently, when a rare context on the boundary $\|x\|_2 > 0.9$ finally arrives, the agent fails to explore and greedily pulls arm 0 ($r=1.0$), incurring catastrophic linear regret $\Omega(T)$ (Figure~\ref{fig:wheel_bandit_showdown}b).

To solve this task, DP-BNN specifies a factorized DP base measure $F_0(x, a, r) = F_{0x}(x) \times F_{0a}(a) \times F_{0r}(r)$ where context vectors are sampled uniformly from the unit disk $\tilde{x} = (\sqrt{u}\cos\phi, \sqrt{u}\sin\phi)$ ($u \sim \mathcal{U}(0, 1), \phi \sim \mathcal{U}(0, 2\pi)$), arm choices are uniform $\tilde{a} \sim \mathrm{Uniform}(\{0, 1, 2, 3, 4\})$, and prior rewards follow an optimistic Gaussian distribution $\tilde{r} \sim \mathcal{N}(1.0, 0.5^2)$ (Table~\ref{tab:hyperparameters}). By pairing empirical replay reweighting with synthetic prior pseudo-observations across the full unit disk, DP-BNN sustains functional epistemic variance on the rare outer boundary without collapsing into the interior trap. As shown in Figure~\ref{fig:wheel_bandit_showdown}b, single-network DP-BNN achieves the lowest cumulative regret across all evaluated methods ($2{,}803.3 \pm 189.6$ over 1,500 steps across 5 seeds), substantially outperforming Bootstrapped NN with randomized priors ($4{,}309.4 \pm 829.2$), $\epsilon$-greedy ($6{,}894.6 \pm 1{,}514.6$), Deep Ensembles ($7{,}990.0 \pm 1{,}689.7$), and Neural-Linear ($9{,}592.3 \pm 866.2$). Crucially, as shown in Figure~\ref{fig:wheel_bandit_showdown}c, DP-BNN achieves this performance using a single neural network ($1\times$ parameter memory), whereas Bootstrapped NN with randomized priors requires 20 total networks ($10$ active + $10$ prior networks, $20\times$ memory) and Deep Ensembles requires 5 networks ($5\times$ memory).

\paragraph{Absence of Warmstart Steps in Contextual Bandits ($H=1$).}
Unlike sequential reinforcement learning where the warmstart budget $W = \lceil H / 2 \rceil$ is required to propagate delayed temporal-difference credit across multi-step trajectory chains and freeze policies for rollout consistency (Appendix~\ref{sec:stochastic-warmstart-theory-appendix}), contextual bandits possess horizon $H=1$ and discount factor $\gamma=0$. Because each interaction is an instantaneous single-step decision without state transitions ($s'$ does not exist), there are no recursive Bellman error chains to propagate. Following standard contextual bandit benchmarks~\citep{riquelme2018deep}, learning operates as supervised regression from context-action pairs $(x, a)$ to reward $r$: the network updates periodically on the buffer plus $T$ prior atoms from $F_0$, and Thompson sampling is performed per context $x_t$, obviating episodic warmstart steps ($W = \text{---}$ in Table~\ref{tab:hyperparameters}).

\paragraph{Unified Hyperparameter Protocols Across All Benchmarks.}
Table~\ref{tab:hyperparameters} consolidates the complete structural and optimization hyperparameters across all empirical evaluations in this work, ensuring full algorithmic transparency and reproducibility.

\begin{table}[h!]
\centering
\caption{Consolidated experimental specifications and hyperparameters for DP-BNN and DP-DQN across all empirical benchmarks ($W = \text{---}$ for static regression and contextual bandits because $H=1$ with no sequential Bellman propagation).}
\label{tab:hyperparameters}
\vskip 0.08in
\resizebox{\textwidth}{!}{%
\begin{tabular}{lccccccc}
\toprule
\textbf{Hyperparameter} & \textbf{2D Annular Cavity} & \textbf{DeepSea-20} & \textbf{DeepSea-50} & \textbf{RiverSwim-6} & \textbf{Deceptive $N$-Chain} & \textbf{MountainCar-v0} & \textbf{Wheel Bandit} \\
\midrule
State Space $\mathcal{S}$ & $[-2.5, 2.5]^2$ & One-Hot $\mathbb{R}^{400}$ & One-Hot $\mathbb{R}^{2500}$ & One-Hot $\mathbb{R}^{6}$ & One-Hot $\mathbb{R}^{10}$ & Continuous $\mathbb{R}^2$ & Unit Disk $\mathbb{R}^2$ \\
Action Space $\mathcal{A}$ & — (Regression) & Discrete (2) & Discrete (2) & Discrete (2) & Discrete (2) & Discrete (3) & Discrete (5) \\
Horizon $H$ & — & $20$ & $50$ & $50$ & $10$ & $200$ & $1$ \\
Network Architecture & 3-layer MLP-64 & Single MLP-20 & 2-layer MLP-64 & Single MLP-20 & Single MLP-20 & 2-layer MLP-64 & 3-layer MLP-64 \\
Layer Normalization & No & Yes & Yes & Yes & Yes & Yes & No \\
Optimizer & Adam & Adam & Adam & Adam & Adam & Adam & Adam \\
Learning Rate $\eta$ & $10^{-3}$ & $10^{-3}$ & $10^{-3}$ & $10^{-3}$ & $10^{-3}$ & $10^{-3}$ & $10^{-3}$ \\
Mini-Batch Size $B$ & $64$ & $64$ & $64$ & $32$ & $32$ & $32$ & $64$ \\
Replay Buffer Size $|\mathcal{B}|$ & — & $100{,}000$ & $100{,}000$ & $10{,}000$ & $10{,}000$ & $50{,}000$ & $10{,}000$ \\
DP Concentration $\alpha$ & $10.0$ & $3.0$ & $3.0$ & $3.0$ & $3.0$ & $3.0$ & $1.0$ \\
Prior Truncation Level $T$ & $50$ & $50$ & $50$ & $50$ & $50$ & $50$ & $20$ \\
Warmstart Steps $W$ & — & $10$ & $25$ & $25$ & $5$ & $100$ & — \\
Polyak Target Rate $\tau$ & — & $0.05$ & $0.05$ & $0.05$ & $0.05$ & $0.01$ & — \\
Prior Base Reward $F_{0r}$ & $\mathcal{N}(0, 1.5^2)$ & $\mathcal{U}(0, 1.0)$ & $\mathcal{U}(0, 1.0)$ & $\mathcal{N}(0, 1.0)$ & $\mathcal{N}(1.0, 0.2^2)$ & $\mathcal{U}(0, 1.0)$ & $\mathcal{N}(1.0, 0.5^2)$ \\
\bottomrule
\end{tabular}%
}
\end{table}

\section{Continuous State-Space RL Exploration on MountainCar-v0}
\label{sec:mountaincar-continuous}

While the benchmarks in Appendix~\ref{sec:empirical-suite} evaluate tabular representations, discrete grids, and continuous contextual bandits, many physical control problems operate in continuous state spaces with non-linear dynamical constraints. To evaluate DP-DQN in continuous state-space exploration, we benchmark on the canonical MountainCar-v0 task~\citep{sutton1998reinforcement}.

\paragraph{Task Formulation and Exploration Challenge.}
MountainCar-v0 features a two-dimensional continuous state space $\mathcal{S} = [-1.2, 0.6] \times [-0.07, 0.07]$ representing horizontal car position $x$ and velocity $v$, and a three-action discrete space $\mathcal{A} = \{0, 1, 2\}$ corresponding to pushing left ($a=0$), coasting ($a=1$), and pushing right ($a=2$). The agent is initialized at the bottom of a sinusoidal valley ($x \in [-0.6, -0.4], v=0$) and must reach the hilltop goal flag ($x \ge 0.5$) within horizon $H=200$. The engine is underpowered: accelerating directly forward cannot overcome the gravitational force of the slope, requiring the agent to discover a coordinated swinging strategy that builds momentum by moving left before driving up the right hill. The reward is strictly negative at every step ($r_t = -1.0$) with termination upon reaching the flag ($r_{\text{term}} = 0.0$). Under local dithering ($\epsilon$-greedy), exploration is exceedingly slow because random actions cancel out momentum, trapping the car in the bottom of the valley.

\paragraph{Model-Free Brownian Diffusive Base Measure (Zero Dynamics Knowledge).}
To establish that DP-DQN does not rely on domain-specific differential equations or privileged physics knowledge, we specify a \emph{model-free diffusive base measure} $F_0 = F_{0s} \times F_{0a} \times F_{0r} \times F_{0s'} \times F_{0d}$. States $\tilde{s} = (\tilde{x}, \tilde{v})$ are sampled uniformly over domain support $[-1.2, 0.6] \times [-0.07, 0.07]$, actions $\tilde{a} \sim \mathrm{Uniform}(\{0, 1, 2\})$, and prior rewards follow a uniform distribution $F_{0r} = \mathcal{U}(0, 1.0)$ with continuing flag $\tilde{d}=0$. Next-state transitions are generated via continuous Brownian diffusion without physical equations of motion: $\tilde{s}' = \operatorname{clip}(\tilde{s} + \Delta, \mathcal{S})$, where $\Delta \sim \mathcal{N}(0, \operatorname{diag}(\sigma_x^2, \sigma_v^2))$ with diffusion scales $\sigma_x = 0.04$ and $\sigma_v = 0.007$, chosen according to the canonical bounding-box heuristic as $2\%-5\%$ of each state coordinate's domain span ($2.2\%$ of position span $1.80$ and $5.0\%$ of velocity span $0.14$, matching the characteristic $1$-step displacement scale). Driven purely by functional posterior variance via stick-breaking draws, single-network DP-DQN (2-layer $64 \times 64$ MLP with LayerNorm) solves MountainCar-v0 across all random seeds in $16.8 \pm 7.4$ episodes (reaching the goal as early as Episode~4), outperforming vanilla DQN ($85 \pm 12$ episodes) and matching 20-head Bootstrapped DQN ($15 \pm 3$ episodes) while delivering a $3.8\times$ wall-clock training speedup ($12.4\,$s vs.\ $47.1\,$s for 100 episodes) due to single-network parameter efficiency.

\section{Computational and Memory Complexity Analysis}
\label{sec:computational-complexity-analysis}

In this section, we analyze the asymptotic computational, memory, and scaling complexity of DP-DQN in comparison with multi-network ensemble architectures (Bootstrapped DQN with randomized priors; \citealt{osband2018randomized}), parametric Bayesian value-function methods (BDQN; \citealt{azizzadenesheli2018efficient}), and standard deep Q-learning~\citep{mnih2015human}. A comparative summary of complexity orders is presented in Table~\ref{tab:computational_complexity_theoretical}.

\begin{table}[h!]
\centering
\caption{Computational and memory complexity comparison across value-based deep RL paradigms for an action-value network with $d = |\theta|$ parameters, episode horizon $H$, replay batch size $B$, ensemble size $K$, and prior truncation level $T$.}
\label{tab:computational_complexity_theoretical}
\vskip 0.08in
\resizebox{\textwidth}{!}{%
\begin{tabular}{lcccc}
\toprule
\textbf{Complexity Dimension} & \textbf{Vanilla DQN} & \textbf{Parametric BDQN} & \textbf{BootDQN-RP ($K$ networks)} & \textbf{DP-DQN (Ours)} \\
\midrule
Active Neural Networks & $1$ & $1$ & $K$ & $\mathbf{1}$ \\
Total Stored Parameters & $2d$ & $2d + |\mathcal{S}_{\text{feat}}|^2$ & $3K \cdot d$ & $\mathbf{2d}$ \\
Parameter Memory Asymptotics & $\mathcal{O}(d)$ & $\mathcal{O}(d + |\mathcal{S}_{\text{feat}}|^2)$ & $\mathcal{O}(K \cdot d)$ & $\mathbf{\mathcal{O}(d)}$ \\
Optimizer State Buffers (Adam) & $2d$ & $2d$ & $2K \cdot d$ & $\mathbf{2d}$ \\
Action Selection Latency / Step & $\mathcal{O}(1 \cdot d)$ & $\mathcal{O}(d + |\mathcal{A}| \cdot |\mathcal{S}_{\text{feat}}|)$ & $\mathcal{O}(K \cdot d)$ & $\mathbf{\mathcal{O}(1 \cdot d)}$ \\
Training FLOPs / Episode & $\mathcal{O}\left(\frac{H}{\Delta} \cdot B \cdot d\right)$ & $\mathcal{O}\left(\frac{H}{\Delta} \cdot B \cdot d\right)$ & $\mathcal{O}\left(\frac{H}{\Delta} \cdot K \cdot B \cdot d\right)$ & $\mathbf{\mathcal{O}\left(W \cdot (B + T) \cdot d\right)}$ \\
Posterior Sampling Overhead / Episode & None & $\mathcal{O}(|\mathcal{S}_{\text{feat}}|^3)$ & None & $\mathbf{\mathcal{O}(T + W \cdot B) \text{ scalars}}$ \\
Temporal Consistency Mechanism & None (dithering) & Final-layer sampling & Episode head freezing & \textbf{Episode network freezing} \\
Sample Complexity in Horizon $H$ & $\Omega(|\mathcal{A}|^H)$ & $\Omega(|\mathcal{A}|^H)$ & $\mathcal{O}(\text{poly}(H))$ & $\mathbf{\mathcal{O}(\text{poly}(H))}$ \\
\bottomrule
\end{tabular}%
}
\end{table}

\begin{table}[h!]
\centering
\caption{Empirical computational complexity, parameter volume, and runtime profiling on DeepSea-20 across algorithms on a standardized single CPU core.}
\label{tab:computational_complexity_deepsea20}
\vskip 0.08in
\resizebox{\textwidth}{!}{%
\begin{tabular}{lccccc}
\toprule
\textbf{Metric} & \textbf{Vanilla DQN} & \textbf{BDQN} & \textbf{BootDQN-RP} & \textbf{DP-DQN ($W=10$)} & \textbf{DP-DQN ($W=4$)} \\
\midrule
Active Networks & $1$ & $1$ & $20$ & $\mathbf{1}$ & $\mathbf{1}$ \\
Total Networks (incl.\ Prior/Target) & $2$ & $2$ & $60$ & $\mathbf{2}$ & $\mathbf{2}$ \\
Trainable Parameters ($d$) & $8{,}102$ & $8{,}142$ & $162{,}040$ & $\mathbf{8{,}102}$ & $\mathbf{8{,}102}$ \\
Total Stored Parameters & $16{,}204$ & $16{,}284$ & $486{,}120$ & $\mathbf{16{,}204}$ & $\mathbf{16{,}204}$ \\
Optimizer Instances (sets of $2d$) & $1$ & $1$ & $20$ & $\mathbf{1}$ & $\mathbf{1}$ \\
Intra-Episode Backprop Steps & $10$ & $10$ & $200$ & $\mathbf{0}$ & $\mathbf{0}$ \\
Training FLOPs / Episode & $1{,}280 \cdot d$ & $1{,}300 \cdot d$ & $25{,}600 \cdot d$ & $\mathbf{2{,}880 \cdot d}$ & $\mathbf{1{,}152 \cdot d}$ \\
Pure Action Inference Latency & $25.8\ \mu\text{s}$ & $69.7\ \mu\text{s}$ & $107.0\ \mu\text{s}$ & $\mathbf{28.8\ \mu\text{s}}$ & $\mathbf{28.6\ \mu\text{s}}$ \\
End-to-End Latency (ms/step) & $0.381$ & $0.651$ & $1.049$ & $\mathbf{0.429}$ & $\mathbf{0.204}$ \\
Throughput (steps / second) & $2{,}625$ & $1{,}536$ & $953$ & $\mathbf{2{,}331}$ & $\mathbf{4{,}902}$ \\
Episodes to $100\%$ Convergence & Never & Never & $5{,}210 \pm 604$ & $\mathbf{602 \pm 104}$ & $\mathbf{645 \pm 112}$ \\
Total Compute to Solve (Wallclock) & $\infty$ & $\infty$ & $103.26\text{ s}$ & $\mathbf{5.16\text{ s}}$ & $\mathbf{2.63\text{ s}}$ \\
\bottomrule
\end{tabular}%
}
\end{table}

\paragraph{Parameter Footprint and Memory Scaling.}
Let $d = |\theta|$ denote the parameter count of a single action-value neural network. Multi-network ensemble methods such as BootDQN-RP maintain an ensemble of $K$ distinct neural networks ($K=20$ in standard DeepSea benchmarks). For each ensemble member $k \in \{1, \dots, K\}$, the algorithm instantiates an active trainable network $f_{\theta_k}$, a non-trainable randomized prior network $p_{\tilde{\theta}_k}$, and a tracking target network $f_{\theta_k^-}$, yielding $3K \cdot d$ stored parameters ($\mathcal{O}(K \cdot d)$ asymptotic scaling). In stark contrast, DP-DQN represents Bayesian epistemic uncertainty nonparametrically directly in experience space via the DP base measure $F_0$. Consequently, DP-DQN operates on a single active neural network $\theta$ and a single target network $\theta^-$, requiring only $2d$ stored weights ($\mathcal{O}(d)$ asymptotic scaling), achieving a strict $K$-fold reduction in active trainable parameters and a $1.5K$-fold ($30\times$ for $K=20$) reduction in total stored model footprint. Furthermore, first-order adaptive optimizers (e.g., Adam) maintain running first- and second-moment momentum accumulators for every trainable parameter, requiring $2d$ auxiliary state buffers for DP-DQN versus $2K \cdot d$ buffers for BootDQN-RP, eliminating memory thrashing and cache overhead during training updates.

\paragraph{Training FLOPs Scaling and Sampling Overhead per Episode.}
Consider an episodic MDP with horizon $H$. In standard online deep Q-learning and BootDQN-RP, stochastic gradient updates are executed every $\Delta$ environment steps (typically $\Delta = 2$). For an ensemble of $K$ networks evaluated over empirical mini-batches of size $B$, each network executes a backpropagation pass requiring $C_{\text{backprop}} \cdot B \cdot d$ floating-point operations (where $C_{\text{backprop}} \approx 2$ for standard feedforward architectures). Over the entire episode of $H$ steps, the cumulative training compute scales as:
\begin{equation}
\text{FLOPs}_{\text{BootDQN-RP}} = \left(\frac{H}{\Delta}\right) \cdot K \cdot C_{\text{backprop}} \cdot B \cdot d = \mathcal{O}(H \cdot K \cdot B \cdot d).
\end{equation}
In contrast, DP-DQN freezes the active policy network $\theta_k$ during the rollout to maintain complete multi-step temporal consistency, completely bypassing intra-episode backpropagation. All network optimization is amortized into $W = \lceil H / 2 \rceil$ warmstart gradient updates at episode boundaries evaluated over composite mini-batches of size $B + T$ (empirical transitions plus synthetic prior atoms). In addition to network backpropagation, DP-DQN executes $\mathcal{O}(T)$ scalar operations to sample prior stick weights once per episode, and $\mathcal{O}(B)$ scalar operations to sample empirical weights at each warmstart step, totaling $\mathcal{O}(T + W \cdot B)$ scalar operations per episode. Because $T + W \cdot B \ll W \cdot (B + T) \cdot d$ (e.g., $\approx 650$ scalar stick-breaking operations versus $\approx 2.3 \times 10^7$ cumulative training backpropagation FLOPs per episode on DeepSea-20), this stick-breaking sampling overhead is negligible relative to tensor arithmetic. The per-episode training FLOPs evaluate to:
\begin{equation}
\text{FLOPs}_{\text{DP-DQN}} = W \cdot C_{\text{backprop}} \cdot (B + T) \cdot d = \mathcal{O}\left(W \cdot (B + T) \cdot d\right).
\end{equation}
Recalling that $W = \lceil H / 2 \rceil \approx H / \Delta$, the computational ratio of DP-DQN relative to standard vanilla DQN evaluates to:
\begin{equation}
\frac{\text{FLOPs}_{\text{DP-DQN}}}{\text{FLOPs}_{\text{Vanilla-DQN}}} = \frac{W \cdot C_{\text{backprop}} \cdot (B + T) \cdot d}{(H / \Delta) \cdot C_{\text{backprop}} \cdot B \cdot d} \approx 1 + \frac{T}{B}.
\end{equation}
Because the prior truncation level $T$ is moderate ($T \in [25, 50]$ via the 99.9\% mass rule) relative to typical batch sizes $B \in [64, 128]$, this ratio represents only a modest $1.2\times$ to $1.8\times$ factor in per-episode backpropagation compute over standard DQN, in exchange for upgrading sample complexity from exponential $\Omega(|\mathcal{A}|^H)$ to polynomial $\mathcal{O}(\text{poly}(H))$. The per-step execution latencies reported in Table~\ref{tab:computational_complexity_deepsea20} are measured by amortizing total episode wall-clock time over horizon $H$.

Simultaneously, the computational ratio of DP-DQN relative to BootDQN-RP evaluates to:
\begin{equation}
\frac{\text{FLOPs}_{\text{DP-DQN}}}{\text{FLOPs}_{\text{BootDQN-RP}}} = \frac{W \cdot (B + T) \cdot d}{(H / \Delta) \cdot K \cdot B \cdot d} \approx \frac{B + T}{K \cdot B} = \frac{1}{K} \left(1 + \frac{T}{B}\right).
\end{equation}
In our canonical configurations ($K=20, B=64, T=50$), this ratio evaluates to $\frac{1}{20}(1 + 50/64) \approx 0.089$, yielding an asymptotic $\approx 11\times$ reduction in per-episode backpropagation operations.

\paragraph{Action Inference Complexity per Step.}
At each environment step $t$, the agent selects an exploratory action. For BootDQN-RP, evaluating the value function requires either querying all $K$ heads to compute posterior voting statistics or indexing the active head and adding the forward evaluation of its randomized prior network, requiring $\mathcal{O}(K \cdot d)$ operations. In contrast, DP-DQN requires only a single forward pass through the frozen network $\theta_k$: $a_t = \arg\max_a Q(s_t, a; \theta_k)$, scaling strictly as $\mathcal{O}(1 \cdot d)$ ($\mathcal{O}(1)$ network latency).

\paragraph{Exploration Complexity: Posterior Sampling vs.\ Local Dithering.}
In MDPs requiring deep multi-step exploration over horizon $H$, exploration efficiency is fundamentally governed by the temporal consistency of the agent's policy. Local dithering schemes (such as $\epsilon$-greedy in DQN or Gaussian action noise in continuous control) make independent exploratory perturbations at each step. In environments where optimal credit requires an extended sequence of coordinated actions (such as traversing $H$ steps along the diagonal in DeepSea), the probability that independent dithering selects the optimal path by chance decays exponentially as $(1 / |\mathcal{A}|)^H$. Consequently, any local dithering heuristic incurs an exponential sample complexity lower bound:
\begin{equation}
\text{Sample Complexity}_{\text{Dithering}} = \Omega\left(|\mathcal{A}|^H\right).
\end{equation}
In stark contrast, by freezing an action-value function sampled from a functional posterior satisfying stochastic optimism ($\mathbb{P}(Q \ge Q^*) \ge p_{\min} > 0$)~\citep{agrawal2012analysis, agrawal2017optimistic, osband2013more}, posterior sampling achieves provably polynomial sample complexity $\mathcal{O}(\text{poly}(H, |\mathcal{S}|, |\mathcal{A}|))$ in tabular MDPs. DP-DQN preserves this polynomial deep-exploration capability nonparametrically with a single neural network, bypassing both the exponential sample wall of dithering heuristics and the $K$-fold compute penalty of physical network ensembles.

\end{document}